\pdfoutput=1
\documentclass[11pt]{article}

\usepackage[final]{acl}

\usepackage{times}
\usepackage{latexsym}

\usepackage[T1]{fontenc}
\usepackage[utf8]{inputenc}

\usepackage{microtype}

\usepackage{inconsolata}

\usepackage{amsmath,amssymb,amsfonts}  % 分别用于多行公式，黑板体字体，其他额外字体
\usepackage{graphicx}  % 处理图像文件必备，例如标题的 LOGO
\usepackage[misc]{ifsym}  % 小信封
\usepackage[hang,flushmargin]{footmisc}  % 禁止脚注缩进
\usepackage{booktabs}  % 支持更好看的表格分割线
\usepackage{tablefootnote}  % 表格中的脚注
\usepackage{tabularx}  % 长表格处理
\usepackage[normalem]{ulem}  % 下划线支持
\useunder{\uline}{\ul}{}  % 下划线支持
\usepackage{algorithm}  % 显示算法
\usepackage{algpseudocode}  % 显示算法
\usepackage[page,header]{appendix}  % 附录的目录
\usepackage{titletoc}  % 附录的目录

\title{
    \includegraphics[scale=0.095]{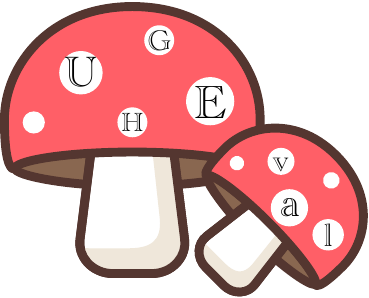}
    $\mathbb{UHG}$Eval: Benchmarking the Hallucination of \\
    Chinese Large Language Models via Unconstrained Generation
    \thanks{
        \textsuperscript{*} These authors contribute equally \\ 
        \textsuperscript{\Letter} Corresponding author: lizy@iaar.ac.cn
    }
}

\makeatletter
\def\thanks#1{\protected@xdef\@thanks{\@thanks\protect\footnotetext{#1}}}
\makeatother

\author{
    \textbf{Xun Liang}\textsuperscript{*}, 
    \textbf{Shichao Song}\textsuperscript{*}, 
    \textbf{Simin Niu}\textsuperscript{*}, 
    \textbf{Zhiyu Li}\textsuperscript{\dag}\textsuperscript{\Letter}, 
    \textbf{Feiyu Xiong}\textsuperscript{\dag}, 
    \textbf{Bo Tang}\textsuperscript{\dag}, 
    \textbf{Yezhaohui Wang}\textsuperscript{\dag}, \\
    \textbf{Dawei He}\textsuperscript{\ddag}, 
    \textbf{Peng Cheng}\textsuperscript{\ddag}, 
    \textbf{Zhonghao Wang}\textsuperscript{\ddag}, 
    \textbf{Haiying Deng}\textsuperscript{\ddag}
    \vspace{.8mm}\\
    \textsuperscript{*}School of Information, Renmin University of China, Beijing, China\\
    \textsuperscript{\dag}Institute for Advanced Algorithms Research, Shanghai, China\\
    \textsuperscript{\ddag}State Key Laboratory of Media Convergence Production Technology and Systems, Beijing, China
}

\begin{document}
\maketitle
\begin{abstract}
Large language models (LLMs) produce hallucinated text, compromising their practical utility in professional contexts. To assess the reliability of LLMs, numerous initiatives have developed benchmark evaluations for hallucination phenomena. However, they often employ constrained generation techniques to produce the evaluation dataset due to cost and time limitations. For instance, this may involve employing directed hallucination induction or deliberately modifying authentic text to generate hallucinations. These are not congruent with the unrestricted text generation demanded by real-world applications. Furthermore, a well-established Chinese-language dataset dedicated to the evaluation of hallucinations is presently lacking. Consequently, we have developed an $\mathbb{U}$nconstrained $\mathbb{H}$allucination $\mathbb{G}$eneration \underline{Eval}uation ($\mathbb{UHG}$Eval) benchmark, containing hallucinations generated by LLMs with minimal restrictions\footnote{Framework, dataset, and results on our project webpage: \url{https://iaar-shanghai.github.io/UHGEval/}.}. Concurrently, we have established a comprehensive benchmark evaluation framework to aid subsequent researchers in undertaking scalable and reproducible experiments. We have also evaluated prominent Chinese LLMs and the GPT series models to derive insights regarding hallucination.
\end{abstract}

\section{Introduction}

\begin{figure}[]
    \centering
    \includegraphics[width=\linewidth]{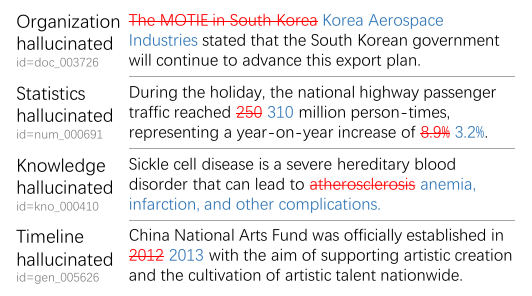}
    \caption{\small
        Hallucinations from $\mathbb{UHG}$Eval. Using the IDs, you can locate the original news articles. 
        \textit{Note}: MOTIE denotes Ministry of Trade, Industry, and Energy.
        (In Chinese: Fig.~\ref{fig:halu_example_ch})
    }
    \label{fig:halu_example}
\end{figure}

% 介绍 LLM 和幻觉
Large language models (LLMs) have unparalleled proficiency in language generation, knowledge application, and intricate reasoning~\cite{LLMSurvey_23_arXiv_RUC}. However, they invariably manifest hallucination~\cite{HalluSurvery2_2023_arXiv_SouthCarolina, yu2024fake}, as they often generate content that is incongruent with user input, the model's output context, or factual information. Real-world hallucination examples from our $\mathbb{UHG}$Eval dataset can be observed in Fig.~\ref{fig:halu_example}.

\begin{figure*}[]
    \centering
    \includegraphics[width=\linewidth]{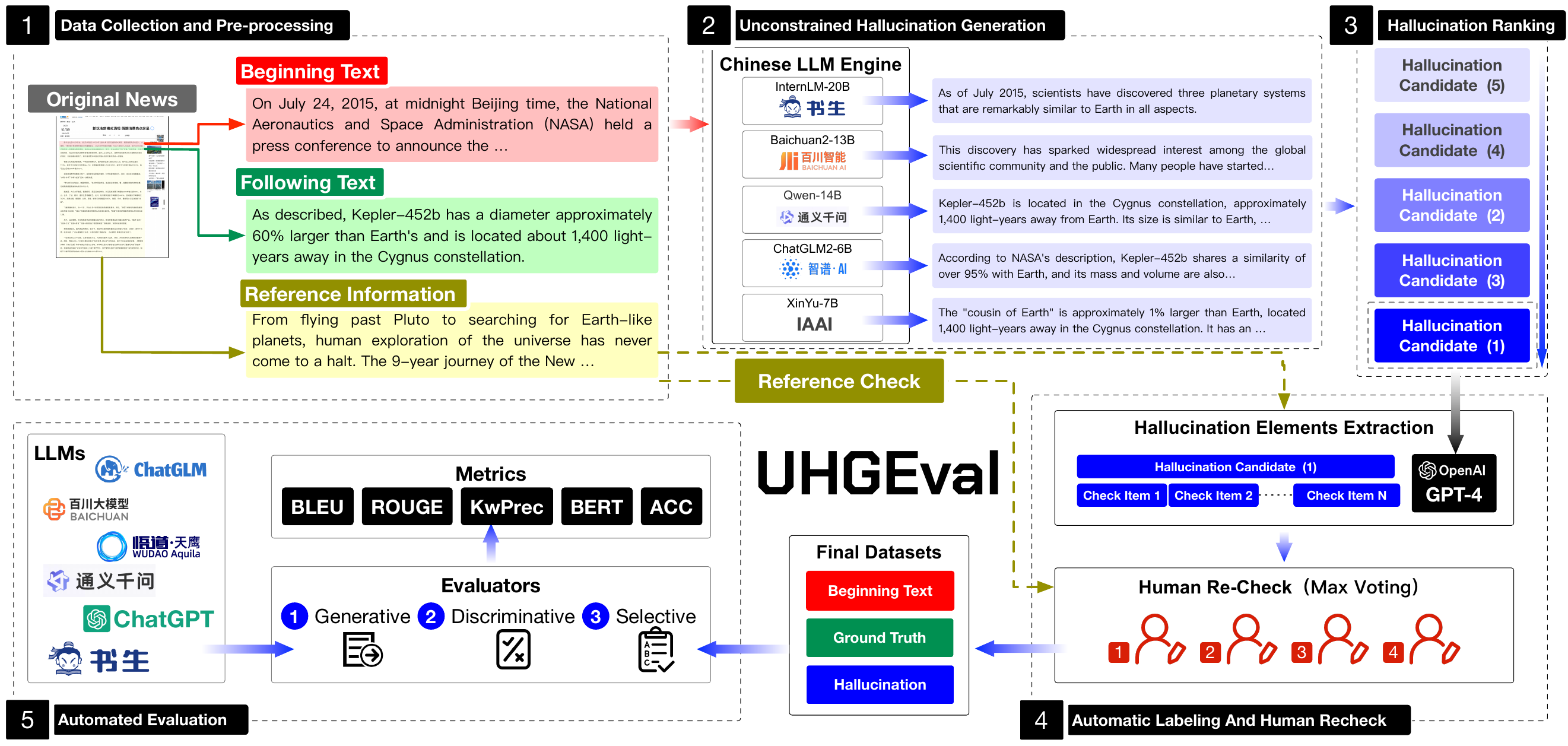}
    \caption{
        \small
        The process of creating $\mathbb{UHG}$Eval. Steps 1 to 4 regarding the creation of the benchmark dataset are explained in Section~\ref{sec:uhgeval}; Step 5, concerning the evaluation framework, is detailed in Section~\ref{sec:experiments}.
        (In Chinese: Fig.~\ref{fig:framework_ch})
    }
    \label{fig:framework}
\end{figure*}

% 幻觉影响落地，亟需高质量基准评测
The fabricated news content depicted in Fig.~\ref{fig:halu_example} offers NO utility to journalists; on the contrary, the verification and rectification of such content exacts a toll on the valuable time of journalists. To this concern, it is crucial to first formulate a comprehensive, stringent, and demanding benchmark for the assessment of hallucination in language generation~\cite{HalluSurvey_23_arXiv_Tencent, HalluSurvery_2023_arXiv_Westlake}.

% 现有 benchmark 有约束生成，且大多数是英语
While there have been a bunch of efforts to develop benchmarks for hallucination assessment, they always employ restricted techniques to produce particular kinds of hallucinated utterances. This approach is at odds with real-world scenarios where hallucinations arise in unrestricted, spontaneously generated content. For example, HaluEval specifies the type of hallucination in the prompt when generating hallucinated text: ``You are trying to answer a question but misunderstand the question context and intention''~\cite{HaluEval_2023_arXiv_RUC}. Additionally, benchmarks such as HaDes annotate hallucinations at a finer granularity by generating token-level hallucinations based on text perturbations~\cite{HADES_2022_ACL_MS}, but the text perturbation method is still constrained. 

% 额外解释无约束生成的重要意义
Hallucinations must be generated in an unconstrained setting; otherwise, it's difficult to determine whether the hallucinated texts in many datasets are indeed errors that language models will make on their own. This point carries profound implications. For example, with a dataset containing freely generated hallucinations, researchers can explore the differences in model hidden states (logits, hidden layers, etc.) between hallucinated text spans and unhallucinated text spans. Such in-depth analysis would not be possible with datasets generated under constrained settings. Appendix~\ref{apdx:comparsions_datasets} provides a detailed comparison with three other datasets, TruthfulQA~\cite{TruthfulQA_2022_ACL_Oxford}, HaluEval~\cite{HaluEval_2023_arXiv_RUC}, and HaDes~\cite{HADES_2022_ACL_MS}.

% 缺少中文评测
Besides, many benchmarks are centered on the evaluation in English, neglecting the assessment of hallucination in Chinese. The extensive lexicon of Chinese characters, combined with the complexities introduced by Chinese word segmentation, renders the Chinese hallucination evaluation particularly arduous and deserving of focused scrutiny.

% 介绍我们的 benchmark
To address the aforementioned challenges, we introduce a novel benchmark for hallucination assessment, as depicted in Fig.~\ref{fig:framework}. The benchmark dataset is composed of raw Chinese news articles and continuations of those articles freely generated by LLMs but annotated with hallucinations.

% 介绍数据领域选择原因
Furthermore, selecting texts from the news domain is intentional, given that news requires utmost precision in conveying factual information and exhibits minimal tolerance for hallucinations, presenting a considerable challenge for the majority of LLMs. Moreover, news data encompasses a wide range of topics, including medicine, technology, finance, sports, etc., incorporating features found in texts from other domains. Lastly, news articles are readily available and frequently employed as training corpora by a large number of LLMs, guaranteeing impartiality in the evaluation of many LLMs~\cite{LLMSurvey_23_arXiv_RUC}.

% 我们的贡献
Our contributions: (1) The development of an unconstrained hallucination evaluation dataset, comprising over 5000 items. Existing methods for constructing datasets often yield biases towards predefined directions, thereby hindering the full simulation of real-world hallucinations. (2) The establishment of a unified and diverse evaluation framework, $\mathbb{UHG}$Eval, that encompasses discriminative, selective, and generative evaluations. Current benchmark methods for hallucination evaluation often exhibit a singular approach and lack task specificity. (3) A comprehensive empirical analysis. We evaluated eight prominent Chinese LLMs and three classic GPT series models to explore the credibility of various LLMs.

\section{Related Work}

% 开头介绍本节要讲的内容
This section outlines hallucination evaluation benchmarks, their characteristics, and evaluation methodologies. A summary of these benchmarks is presented in Table~\ref{tab:benchmarks}. For related works on LLMs and hallucinations, please refer to Appendix~\ref{apdx:related}.

\begin{table*}[h]
\fontsize{6.7}{9}\selectfont
\centering

\begin{tabular}{llllll}
\toprule
Benchmark                                            & Generation Method: Base Dataset  & Annotation   & Metric                       & Granularity       & Lang. \\
\midrule
ChineseFactEval~\cite{ChineseFactEval_2023_web_SJTU} & Manual                           & Manual       & Acc                          & Sentence          & CN    \\
CSK-PN~\cite{CSK-PN_2023_ACL_Fudan}                  & Direct: Common KGs               & No Need      & Acc                          & Word              & EN    \\
FACTOR~\cite{FACTOR_2023_arXiv_AI21Lab}              & CHG: Wiki, News                  & Auto         & FACTOR Acc                   & Sentence          & EN    \\
FActScore~\cite{FActScore_2023_arXiv_UWashington}    & CHG: Wiki                        & No Need      & FActScore by Human           & Short Sentence    & EN    \\
FactualityPrompts~\cite{Factuality_2022_NIPS_HKUST}  & Direct: Wiki                     & Auto         & NE Error, Entailment         & Document, Sentence& EN    \\
HaDes~\cite{HADES_2022_ACL_MS}                       & CHG: Wiki                        & Manual       & Acc, G-Mean, BSS, AUC, etc.  & Word              & EN    \\
HalluQA~\cite{HalluQA_2023_arXiv_Fudan}              & CHG, Manual: TruthfulQA, Wiki    & Manual, Auto & Non-hallucination Rate       & Sentence          & CN    \\
HaLoCheck~\cite{HaLo_2023_arXiv_UPittsburgh}         & CHG                              & No Need      & HaLoCheck, selfcheckGPT      & Sentence          & EN    \\
HaluEval~\cite{HaluEval_2023_arXiv_RUC}              & CHG: Alpaca, HotpotQA, etc.      & Manual, Auto & Acc                          & Document          & EN    \\
HILT~\cite{HalluSurvery2_2023_arXiv_SouthCarolina}   & CHG: NYT, Politifact             & Manual       & HVI                          & Word              & EN    \\
KoLA-KC~\cite{KoLA_2023_arXiv_THU}                   & Direct: Wiki, evolving dataset   & Auto         & BLEU, ROUGE                  & Document          & EN    \\
Med-HALT~\cite{Med-HALT_2023_arXiv_SaamaIndia}       & Direct: MedMCQA, PubMed, etc.    & No Need      & Acc, Pointwise Score         & All               & EN    \\
PHD~\cite{PHD_2023_arXiv_PKU}                        & CHG: Wiki                        & Manual       & F1, Acc, Prec, Reca          & Document          & EN    \\
SelfAware~\cite{SelfAware_2023_ACL_NUS}              & CHG: Quora, HowStuffWorks        & Manual       & F1, Acc                      & Sentence          & EN    \\
STSN~\cite{STSN_2023_arXiv_ArizonaStateU}            & UHG                              & Manual       & Acc, Prec, Reca              & Sentence, Concept & EN    \\
TruthfulQA~\cite{TruthfulQA_2022_ACL_Oxford}         & Manual                           & Manual       & Acc by Human or GPT-judge    & Sentence          & EN    \\
$\mathbb{UHG}$Eval (Ours)                            & UHG: News                        & Auto, Manual & Acc, kwPrec, BERTScore, etc. & Sentence, Keyword & CN    \\
XSum Hallu~\cite{XSum_2020_ACL_Google}               & UHG: XSum                        & Manual       & ROUGE, BERTScore, Acc, etc.  & Word, Document    & EN    \\
\bottomrule
\end{tabular}

\caption{\small Hallucination evaluation benchmarks sorted by name. In the Generation Method column, CHG refers to constrained hallucination generation, UHG refers to unconstrained hallucination generation, Manual indicates manually constructed, and Direct implies utilizing the base dataset without the need for generation. In the Annotation column, Auto denotes automatic machine annotation. 
In the Metric column, Acc, Prec, and Reca respectively indicate accuracy, precision, and recall.
In the Lang. column, CN and EN respectively stand for Chinese and English.}
\label{tab:benchmarks}

\end{table*}

\subsection{Benchmark Dataset Construction}

% 数据集构建
Dataset construction usually involves three steps. Firstly, real-world texts for hallucination generation are collected, and most benchmarks directly use existing datasets, such as Wiki~\cite{FACTOR_2023_arXiv_AI21Lab}, Alpaca~\cite{HaluEval_2023_arXiv_RUC}, PubMed~\cite{Med-HALT_2023_arXiv_SaamaIndia}, etc. Secondly, hallucinations are generated usually by LLMs such as GPT3.5-Turbo, and most works use a constrained hallucination generation (CHG) paradigm. STSN~\cite{STSN_2023_arXiv_ArizonaStateU} and XSum Hallu~\cite{XSum_2020_ACL_Google} are the only two benchmarks that use UHG as we do. Thirdly, it is not certain that the content generated by the LLMs actually contains hallucinations, and often requires annotation, which is mostly done by human involvement. There are also works using automatic machine labeling~\cite{FACTOR_2023_arXiv_AI21Lab, Factuality_2022_NIPS_HKUST, HalluQA_2023_arXiv_Fudan}. These are the basic methods for constructing datasets, but there are also some other paradigms, such as constructing the dataset purely using manual labor, e.g. ChineseFactEval~\cite{ChineseFactEval_2023_web_SJTU}, HaDes~\cite{HADES_2022_ACL_MS}, TruthfulQA~\cite{TruthfulQA_2022_ACL_Oxford}, etc.

\subsection{Benchmark Dataset Characteristics}

% 数据集特征：粒度，语言
Regarding the granularity of hallucinations labeled in the datasets, most studies assess hallucinations at the sentence and document levels, while a few examine them at the word (or keyword, concept) level. Concerning language, most evaluation datasets are in English. To our knowledge, the only two Chinese benchmarks, ChineseFactEval~\cite{ChineseFactEval_2023_web_SJTU} and HalluQA~\cite{HalluQA_2023_arXiv_Fudan} contain only 125 and 450 questions, respectively. Given the notably limited size of these datasets, our work significantly enhances the pool of data available for Chinese hallucination evaluation.

\subsection{Evaluation Schemes}

% 评测方法
Currently, building automatic metrics for evaluation is still dominant, and a small proportion of works use human evaluation~\cite{FActScore_2023_arXiv_UWashington, TruthfulQA_2022_ACL_Oxford, XSum_2020_ACL_Google}. In terms of specific evaluation metrics, most works adopt common classification metrics, e.g., F1, accuracy, precision, and recall. Some other works construct their calculation methods, e.g., FACTOR~\cite{FACTOR_2023_arXiv_AI21Lab}, FActScore~\cite{FActScore_2023_arXiv_UWashington}, HaLoCheck~\cite{HaLo_2023_arXiv_UPittsburgh}, etc. However, the above metrics are rule-based and can only evaluate the ability of LLMs to classify hallucinations, but not the ability of LLMs to generate content without hallucinations. Thus, some benchmarks explore further in generative evaluation. For example, KoLA~\cite{KoLA_2023_arXiv_THU} evaluates knowledge creation (KC) using BLEU and ROUGE, and TruthfulQA~\cite{TruthfulQA_2022_ACL_Oxford} evaluates hallucinations using a specially trained classifier, GPT-judge.

\section{The UHGEval Dataset} \label{sec:uhgeval}

\subsection{Data Collection and Pre-processing}

% 数据选取及原因
We amassed tens of thousands of historical news articles from leading Chinese news websites, covering the period from January 2015 to January 2017, to serve as the foundation for constructing the dataset. It is worth noting that the decision to eschew more recent news articles (e.g., from 2024) was made to better assess the model's understanding of existing knowledge. Indeed, the knowledge embedded within the training data of existing Chinese LLMs typically encompasses information about significant news between 2015 and 2017~\cite{LLMSurvey_23_arXiv_RUC}.

% 类别划分
The collected news spans various topics, such as sports, education, science, society, finance, and more. This diversity underscores the advantage of choosing news texts for our dataset, as it enables the incorporation of a wide array of text genres. We hypothesize that the occurrence of hallucinations will vary as LLMs generate news across different categories. As a result, we have classified these diverse categories into four main types: document-intensive, number-intensive, knowledge-intensive, and general news, with details provided in Table~\ref{tab:collected_news}.

% Considering the different categories of news, such as sports, education, science, and society, the generated hallucinations typically exhibit certain differences. Therefore, when curating the initial news collection for continuation, we endeavored to ensure that the distribution of the collection aligns with the original distribution by randomly sampling from the entire news dataset. Furthermore, we have categorized the collected news examples into four major types: document-intensive, number-intensive, knowledge-intensive, and general news, as shown in Table~\ref{tab:collected_news}. 
% We hypothesize that the likelihood of generating hallucinations varies for different types of news. For example, number-intensive news frequently contains various numerical data, such as years, scores, and values, which may predispose the model to fabricating numbers or introducing minor deviations. Document-intensive news, on the other hand, primarily references official documents, such as factual policy documents, official statements, standard explanations, and legal clauses. In this case, the model may be inclined to fabricate specific policy or document names or create detailed but fictional policy content. Knowledge-intensive news is characterized by an emphasis on enduring truths and analytical reasoning, which can render the model susceptible to flawed reasoning or the retrieval of incorrect facts. In addition to these three types, we also categorize culturally relevant general news as a separate category for experimental control.
% TODO：一个关于四种类别划分的假设，未来可能可以用上。

\begin{table}[]
    \small
    \centering
    \begin{tabularx}{\columnwidth}{p{0.04\textwidth}p{0.05\textwidth}p{0.3\textwidth}}
        \toprule
        \textbf{Type} & \textbf{Share} & \textbf{Categories} \\
        \midrule
        DOC & 27.52\% & Politics, Law, Military, Education \\
        \specialrule{0em}{3pt}{0pt}
        NUM & 43.34\% & Sports, Economy, Market \\
        \specialrule{0em}{3pt}{0pt}
        KNO &  6.55\% & Science, Technology, Healthcare \\
        \specialrule{0em}{3pt}{0pt}
        GEN & 22.59\% & Society, Culture, Arts, Entertainment, Weather, Environmental Protection, Disasters, Accidents \\
        \bottomrule
    \end{tabularx}
    \caption{\small Statistics of collected news. DOC, NUM, KNO, and GEN denote document-intensive, number-intensive, knowledge-intensive, and general news, respectively.}
    \label{tab:collected_news}
\end{table}

% 一篇文章被划分为三个部分
In the data pre-processing stage, we divide a complete news article into three parts: the beginning text, the following text, and the reference information. The beginning text serves to guide the model in generating the continuation and is typically the opening portion of the news. During evaluation, the LLM is required to generate content following the beginning text. The following text comprises the subsequent sentences in the news article and serves as the ground truth for the continuation task. Finally, all the remaining text, after the beginning text is excluded, serves as a source of reference information. This section provides reference information for labeling and also acts as the reference text for the reference-based evaluation.

% % 筛选设置
% \paragraph{Filtering Settings} To ensure the overall quality of the final evaluation dataset, we have implemented the following filters: We consider only the categories listed in Table~\ref{tab:collected_news}, which correspond to the most frequently occurring categories in the original news collection. For news length, we set parameters such that the body length of the selected news falls between 630 and 870 characters, while the beginning text spans between 80 and 120 characters and consists of 2 to 5 sentences. These length parameters reflect the average values in the original news collection and were chosen to avoid overburdening the annotation process at a later stage.
% TODO 暂时不需要，且在 UHGEval-dataset 中有具体描述

\subsection{Unconstrained Hallucination Generation}

% 提示无约束
Unlike directed hallucination generation~\cite{HaluEval_2023_arXiv_RUC} or perturbation-based generation~\cite{HADES_2022_ACL_MS},
we have adopted an unconstrained generation methodology for the continuation of natural language content, though it poses difficulties for subsequent annotations. This generation's fashion entails directly inputting the text to be continued into the model without any restrictive prompt instructions, thereby obtaining organic results.

% 大家使用单个模型生成，我们使用多个模型生成
Furthermore, current benchmarks for evaluating hallucination have predominantly relied on a single LLM to produce a hallucinated dataset. Notable examples include HaluEval~\cite{HaluEval_2023_arXiv_RUC} and PHD~\cite{PHD_2023_arXiv_PKU}, which exclusively utilize ChatGPT, and FActScore~\cite{FActScore_2023_arXiv_UWashington} and FACTOR~\cite{FACTOR_2023_arXiv_AI21Lab}, which solely employ InstructGPT~\cite{InstructGPT_2022_NeurIPS_OpenAI}. In contrast, our methodology incorporates a suite of five distinct Chinese LLMs to generate hallucinated content. These models include ChatGLM2-6B~\cite{GLM_22_ACL_CCFA}, Baichuan2-13B~\cite{Baichuan2_23_arXiv_Baichuan}, Qwen-14B~\cite{Qwen_23_arXiv_Alibaba}, InternLM-20B~\cite{InternLM_23_GitHub_InternLMTeam}, and Xinyu-7B. For additional information about the Xinyu series models, please refer to the Appendix~\ref{apdx:models}.

% 总结
For each input news article, we concurrently generate five candidate continuations using five different LLMs without constraint. Overall, our approach engenders a more unconstrained and heterogeneous generation of hallucinations, mitigating the bias that may arise from the use of a single model or constrained prompting.

\subsection{Hallucination Ranking} \label{sec:hallurank}

% 挑战：UHG生成带来的成本和准确性之间的矛盾
Given the unconstrained nature of our paradigm, the task of discerning whether the generated content is indeed hallucinated presents a significant challenge. Upon generating the continuations, an exclusive dependence on human annotation would incur substantial costs, whereas a purely machine-based approach, such as utilizing GPT4, could potentially yield less accurate results.

% 解决挑战的办法：两阶段标注法，以及各阶段的作用（主要是为了减轻人工标注压力）
To navigate these complexities, we have adopted a two-stage annotation. This approach begins with an initial stage of hallucination ranking (Section \ref{sec:hallurank}), designed to sort the generated content based on the likelihood of hallucination. The ranking is then followed by the second stage of automatic labeling and human rechecking (Section \ref{sec:autolabel}).

% Hallucination Ranking 为什么有必要，算法整体框架和最终结构
Hallucination ranking is a crucial step in selecting the most appropriate continuation from the five candidates generated by the five LLMs. This process relies on two critical metrics: $fluency$, ensuring that the continuation does not become too nonsensical, and $likelihood$, which stands for the likelihood of hallucination occurrence, ensuring that the continuation includes a detectable level of hallucinations. They are computed as follows.

% 流畅度维度
\paragraph{Fluency} This refers to the coherence and readability of the text~\cite{liang2024controlled}. A fluent text should read smoothly and be grammatically correct in the context of the continuation. To assess fluency, a reward model developed by the Institute for Advanced Algorithms Research (IAAR) is employed, trained to score text fluency. The model is fine-tuned using a dataset annotated with news on an open-source reward model, Ziya model\footnote{\url{https://huggingface.co/IDEA-CCNL/Ziya-LLaMA-7B-Reward}}.

\begin{figure}[t]
    \centering
    \includegraphics[width=\linewidth]{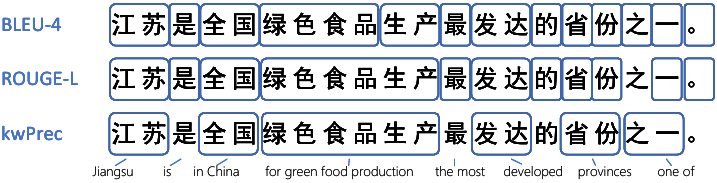}
    \caption{\small Tokenization results for BLEU-4, ROUGE-L, and kwPrec, using $\text{news}_\text{id=num\_000432}$ as an example. The meaning of the above sentence: Jiangsu is one of the most developed provinces in China for green food production.}
    \label{fig:tokenize}
\end{figure}

% 幻觉发生可能性维度
\paragraph{Likelihood of Hallucination Occurrence} This dimension evaluates the extent to which the continuation may contain hallucinated content. To estimate the probability, we evaluate the lexical correlation between the generated continuation and the reference information. The lower the correlation, the more likely hallucinations are to occur. Despite existing metrics, such as BLEU~\cite{BLEU_2002_ACL_WatsonResearch} and ROUGE~\cite{ROUGE_2004_ACL_USouthernCalifornia}, we believe that these rule-based methods may not effectively discover hallucinations. Therefore, we propose the keyword precision (kwPrec) metric.

% 介绍kwPrec
kwPrec uses an LLM (e.g., GPT3.5-Turbo) to extract keywords from the continuation and then determine whether these keywords have exact matches in the reference information. The ratio of all matches to the total keywords is then calculated. Since LLMs often extract appropriate keywords more effectively, kwPrec focuses more on factual relevance rather than expressional relevance. Fig.~\ref{fig:tokenize} illustrates the tokens segmented by kwPrec compared to those obtained by BLEU-4 and ROUGE-L. The prompt template utilized for extracting keywords is depicted in Fig.~\ref{fig:templates0} within Appendix~\ref{apdx:tempalte}.

\begin{algorithm}
\caption{Hallucination Ranking}
\label{algo:rank}
\begin{algorithmic}
\Require $candidate: list[str]$
\Ensure $final: str$

\State $candidate.sort(descend, by=fluency)$
\State $picked \gets candidate[:3]$
\Comment{More fluency} \\

\State $picked.sort(ascend, by=kwPrec)$
\State $final \gets picked[0]$
\Comment{More Hallucination}
\end{algorithmic}
\end{algorithm}

% 重述问题目标
With $fluency$ and $kwPrec$, our task in the hallucination ranking step is to select one out of five candidate continuations that appears to be correct (highest in $fluency$) but is likely to contain hallucinations (lowest in $kwPrec$).

% 具体步骤
The specific steps are as follows (also shown in Algorithm~\ref{algo:rank}). Step 1: Rank the five candidate continuations in descending order by $fluency$. Step 2: Select the top three continuations with the highest $fluency$. Step 3: Rank these three continuations in ascending order by $kwPrec$. Step 4: Choose the continuation with the lowest $kwPrec$ score. Following these steps, the continuation selected in Step 4 is the $final$ choice. By employing such a ranking, it is guaranteed that, in the worst-case scenario, the $final$ candidate ranks at least third in fluency and third in the likelihood of hallucination occurrence, achieving a balanced level.

\subsection{Automatic Labeling and Human Rechecking} \label{sec:autolabel}

\begin{figure}[h]
    \centering
    \includegraphics[width=\linewidth]{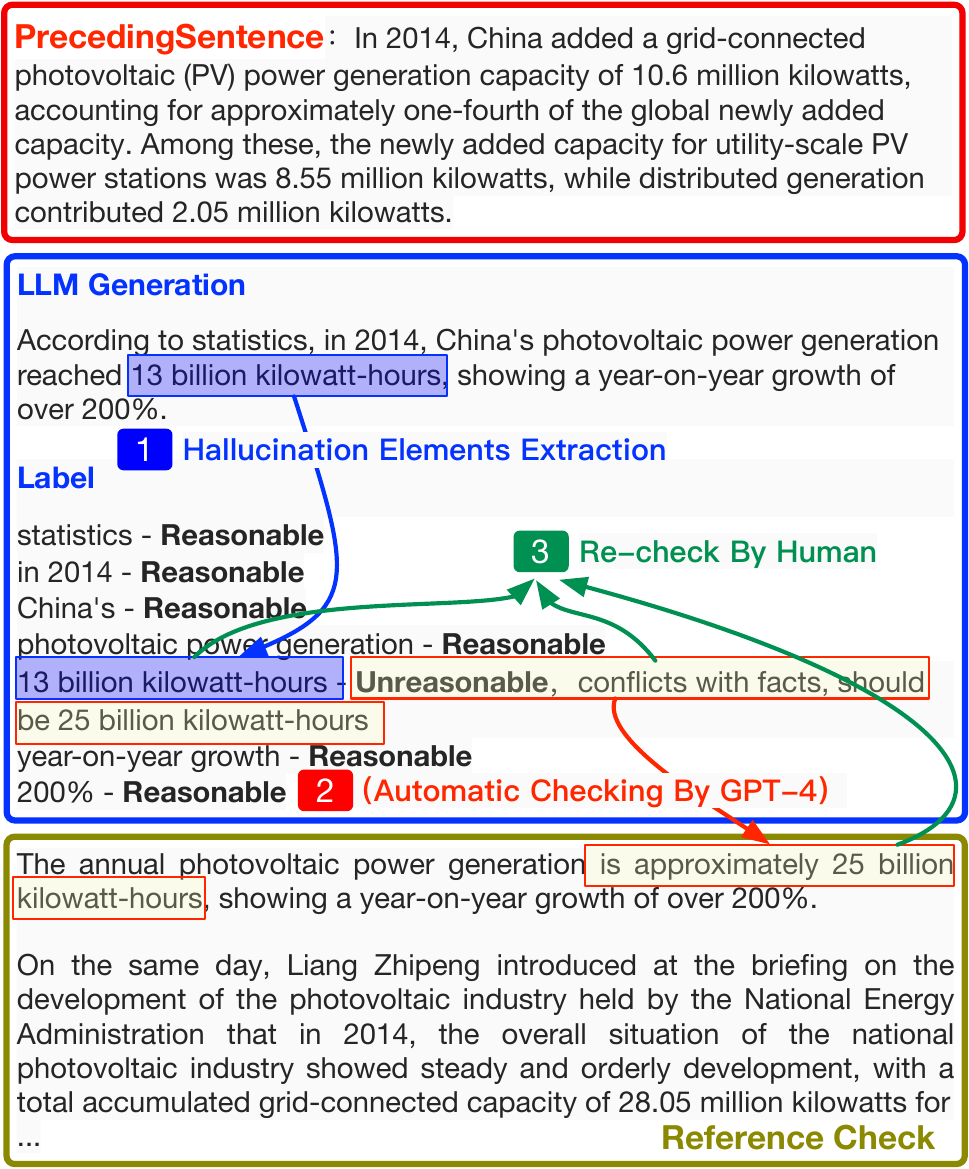}
    \caption{
        \small 
         Labeling and rechecking.
        (In Chinese: Fig.~\ref{fig:annotation_ch})
    }
    \label{fig:annotation}
\end{figure}

% 我们提出的标注方案
Through hallucination ranking, we can identify continuations that are both articulately expressed and likely to contain hallucinations. To detect continuations with confirmed hallucinations, we propose an annotation scheme that utilizes keywords, which includes automatic labeling and subsequent human verification, as shown in Fig.~\ref{fig:annotation}.

% 自动标注
\paragraph{Automatic labeling} We utilize the keywords identified by GPT3.5-Turbo from the candidate continuations, similarly to the process used in the computation of kwPrec previously. These keywords act as the focal points for subsequent verification. Thereafter, we employ GPT4-0613~\cite{GPT4_23_arXiv_OpenAI} to perform annotation on these keywords. It evaluates the validity of the keywords in the continuations by conducting a cross-reference with the provided original news and provides explanations for any detected unreasonable keywords.

% 人工复检
\paragraph{Human rechecking} We undertake a manual, one-to-one verification process by analyzing the annotated results and explanations provided by GPT4-0613 against the original news. This step ensures the accuracy of the machine-generated annotations. In the end, instances verified as accurate by annotators comprise the final $\mathbb{UHG}$Eval dataset. For details on manual annotation, refer to Appendix~\ref{apdx:human}.

\subsection{Dataset Statistics}

% 数据集统计
Starting with 17,714 candidate hallucinated continuations, we curated a dataset of 5,141 hallucinated continuations, as detailed in the basic statistics in Table~\ref{tab:dataset}. For further analysis, the data volume of each step in the dataset creation pipeline, and an example of the dataset, please refer to Appendix~\ref{apdx:data}, Appendix~\ref{apdx:data_volume} and Appendix~\ref{apdx:an_example}, respectively.

\begin{table}[h]
    \small
    \centering
    \begin{tabularx}{\columnwidth}{p{0.14\textwidth}p{0.05\textwidth}p{0.05\textwidth}p{0.05\textwidth}p{0.05\textwidth}}
        \toprule
                             & DOC     & KNO     & NUM     & GEN     \\
        \midrule
        \#news               & 1242    & 320     & 2431    & 1148    \\
        avg. \#hallu. kw.    & 2.15    & 1.99    & 2.54    & 2.12    \\
        avg. \#kw.           & 8.43    & 8.09    & 8.07    & 8.17    \\
        \#hallu. kw. / \#kw. & 25.47\% & 24.61\% & 31.44\% & 26.00\% \\
        avg. len. contn.     & 46.77   & 48.36   & 44.47   & 45.97   \\
        avg. len. begin.     & 102.15  & 102.66  & 103.20  & 102.86  \\
        avg. len. refer.     & 634.17  & 618.90  & 624.47  & 632.47  \\
        \bottomrule
    \end{tabularx}
    \caption{\small Dataset basic statistics. \# denotes quantity, avg. denotes average, len. denotes length, contn. denotes hallucinated continuations, begin. denotes news beginnings, and refer. denotes reference information.}
    \label{tab:dataset}
\end{table}

\section{Experiments} \label{sec:experiments}

\subsection{Models}

% 模型选择总述
Given that our dataset is tailored for the Chinese language generation domain, we selected eight widely used Chinese LLMs and three LLMs from OpenAI. These LLMs are from eight base models: Aquila2~\cite{Aquila2_23_GitHub_BAAI}, Baichuan2~\cite{Baichuan2_23_arXiv_Baichuan}, GLM~\cite{GLM_22_ACL_CCFA}, GPT\footnote{\url{https://openai.com}}, InternLM~\cite{InternLM_23_GitHub_InternLMTeam}, Qwen~\cite{Qwen_23_arXiv_Alibaba}, BLOOMZ~\cite{BLOOMZ_2023_arXiv_HF}, and LLaMA2~\cite{LLaMA2_2023_arXiv_Meta}. Refer to the Appendix~\ref{apdx:models} for a detailed overview of the LLMs used in the experiments.

\subsection{Evaluation Forms}

% 评测的三个维度，引入附录
In this study, we conducted a detailed analysis of evaluation methods across three dimensions: form, metric, and granularity. A more comprehensive report can be found in the Appendix~\ref{apdx:eval}. Here, we introduce the three forms of evaluation.

Firstly, there is the discriminative evaluation, which involves having the model determine whether a continuation contains hallucinations. Secondly, similar to discriminative evaluation, selective evaluation allows LLMs to choose the continuation without hallucinations from options with and without such content. Lastly, we have generative evaluation. Specifically, the LLM under evaluation is provided with a beginning text and is then tasked with generating a continuation. Subsequently, various reference-based techniques are employed to assess whether the generated continuation includes hallucinations.

\subsection{Evaluation Framework}

\begin{figure}[t]
    \centering
    \includegraphics[width=\linewidth]{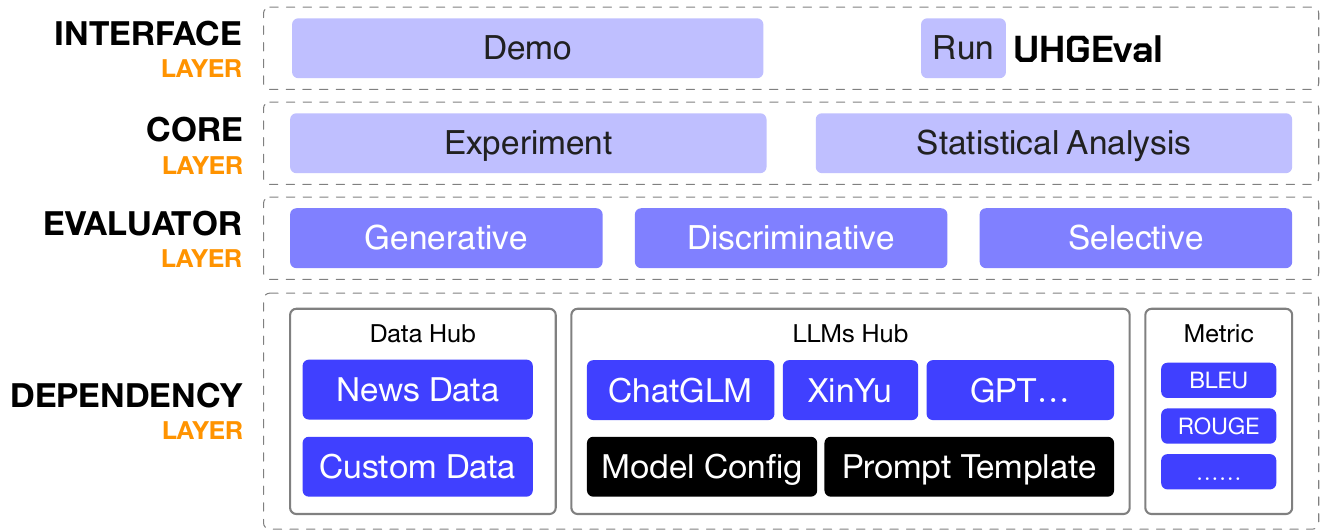}
    \caption{\small Evaluation framework}
    \label{fig:eval_framework}
\end{figure}

% 框架简介
To accommodate different forms of evaluation methods, we have developed a data-secure, easy-to-extend, and easy-to-use evaluation framework, as illustrated in Fig.~\ref{fig:eval_framework}. 
Refer to Appendix~\ref{apdx:framework} for a more detailed understanding of the various layers of the framework.

% 两大特征：对用户来说安全和易用；对开发者来说灵活
$\mathbb{UHG}$Eval is both intuitive and secure for users, offering efficient usage while concurrently ensuring the integrity of experimental results through robust resistance to exceptions and support for resuming evaluations post unexpected interruptions. For developers and researchers, the modules within the Dependency and Evaluator layers are fully interchangeable, thereby affording considerable flexibility for expansion.

\subsection{Experimental Setup}

% % 总括
% To establish a robust experimental framework, we have set up some configurations as follows.

% 提示工程
\paragraph{Prompt Engineering} We apply the technique of ``intent + instruction + 3-shot (explainable) prompting.'' Intent delineates the role, instruction outlines the task, and the prompt incorporates three examples to aid the few-shot learning~\cite{chen2024grimoire,yu2024xfinder}. 
Furthermore, political content in examples is prohibited to adhere to content policies from model providers. 
Explainable prompting entails not merely acquiring results, but also eliciting the model's rationale behind its responses. Refer to Appendix~\ref{apdx:tempalte} to view the complete templates.

% 正负样例平衡
\paragraph{Example Balancing} To guarantee the reliability of experimental outcomes for all LLMs, we meticulously balance examples in discriminative and also in selective evaluations. Specifically, the LLM under evaluation will encounter an equal number of examples with and without hallucinations. % This approach addresses the tendency of some models to learn patterns from the three examples in the prompts and produce conjectural rather than reasoned responses when making judgments. Such a tendency can introduce a considerable bias towards certain outcomes.

% 超参数设置
\paragraph{Hyperparameter Settings} Managing parameters for heterogeneous LLMs is a multifaceted endeavor, as different LLMs feature unique interface designs, and the same parameters can have varying implications across LLMs. 
% For example, the level of determinism influenced by the temperature parameter varies. 
Despite these challenges, we commit to the principle of ``guaranteeing overall output determinism while allowing for slight randomness, and aiming for consistent parameter settings across models.'' Consequently, we set the temperature to 0.1, the top\_p to 0.9, the top\_k to 5, and the random seed to 22.
%Consequently, we configured parameters including temperature, top\_p, top\_k~\cite{LLMSurvey_23_arXiv_RUC}, and random seed. To ensure output determinism and improve reproducibility, we set the temperature to 0.1. Considering that OpenAI models advise against adjusting temperature and top\_p simultaneously, we minimally altered top\_p, setting it at 0.9. We set top\_k to 5, which is effective for certain models. To further enhance reproducibility, we established a seed for random number generators, setting it at 22.

% 指标设置
\paragraph{Metrics}
For discriminative and selective evaluation, accuracy serves as the metric. For generative evaluation, metrics consist of 4-gram BLEU (BLEU-4), the longest common subsequence-based ROUGE (ROUGE-L), kwPrec, and BERTScore.

\subsection{Results and Analysis}

Results are presented in Table~\ref{tab:results} and Table~\ref{tab:eval_by_type}.

\begin{table*}[h!]
    \small
    \centering
    \begin{tabularx}{\textwidth}{p{0.16\textwidth}p{0.09\textwidth}p{0.1\textwidth}p{0.09\textwidth}p{0.09\textwidth}p{0.08\textwidth}p{0.08\textwidth}p{0.08\textwidth}}
        \toprule
                      & \multicolumn{3}{c}{\textbf{Discriminative-Keyword}} & \multicolumn{2}{c}{\textbf{Discriminative-Sentence}} & \multicolumn{2}{c}{\textbf{Selective}} \\
                      \cmidrule(lr){2-4} \cmidrule(lr){5-6} \cmidrule(lr){7-8}
                      & avg. acc. & avg. \#kws & \#valid & avg. acc.         & \#valid         & acc.           & \#valid      \\
        \midrule
        Aquila-34B    & 53.62\%   & 3.00       & 3719    & 49.86\%           & 5009            & 54.29\%       & 4319          \\
        Baichuan2-13B  & 51.63\%   & \textbf{3.128}       & 4478    & 46.88\%           & 5047            & 50.23\%       & 5130          \\
        Baichuan2-53B & 52.13\%   & 2.98       & 1656    & 50.81\%           & 1478            & 54.67\%       & 4443          \\
        ChatGLM2-6B   & 50.80\%   & 3.10       & 4289    & 43.87\%           & 5130            & 43.59\%       & 5130          \\
        GPT3.5-Turbo & 53.72\%   & 3.08       & 4183    & 50.02\%           & 5039            & 49.03\%       & 5103          \\
        GPT4-0613    & \textbf{70.04\%}   & 3.07       & 4100    & \textbf{57.42\%}           & 5024            & 55.20\%       & 5047          \\
        GPT4-1106     & {\ul 69.48\%}   &  3.10   & 4189    & {\ul 57.38\% }           & 4903            &  \textbf{60.35\%}      & 4752  \\
        InternLM-20B  & 50.92\%   & 3.10       & 4388    & 51.01\%           & 5130            & 49.43\%       & 5130          \\
        Qwen-14B      & 52.86\%   & {\ul 3.125}       & 4478    & 50.58\%           & 5130            & 54.74\%       & 5130          \\
        Xinyu-7B      & 49.58\%   & 3.12       & 4451    & 48.66\%           & 5014            & 50.58\%       & 5130          \\
        Xinyu2-70B     & 52.94\%   & 3.12       & 4482    & 55.04\%          & 5128            & {\ul 57.93\%}       & 5129          \\
        \bottomrule
    \end{tabularx}

    \vspace{5mm}
    
    \begin{tabularx}{\textwidth}{p{0.16\textwidth}p{0.111\textwidth}p{0.111\textwidth}p{0.111\textwidth}p{0.111\textwidth}p{0.111\textwidth}p{0.1\textwidth}}
    \toprule
                  & \multicolumn{6}{c}{\textbf{Generative}} \\
                  \cmidrule(lr){2-7}
                  & avg. bleu & avg. rouge & avg. kwPrec & avg. bert & avg. len. & \#valid \\
    \midrule
    Aquila-34B    & 11.80\%   & 6.04\%     & \textbf{34.36\%}     & 67.51\%   & 43.76    & 5130    \\
    Baichuan2-13B & 8.84\%    & 6.96\%     & 25.51\%     & 65.69\%   & 46.04    & 5113    \\
    Baichuan2-53B & 10.06\%   & {\ul 7.55\%}     & 26.45\%     & 67.65\%   & 49.40    & 3837    \\
    ChatGLM2-6B   & 9.17\%    & 7.17\%     & 24.53\%     & 64.89\%   & 46.27    & 5094    \\
    GPT3.5-Turbo & 9.02\%    & 6.30\%     & 27.74\%     & 66.39\%   & 39.04    & 5084    \\
    GPT4-0613    & 10.74\%   & 7.19\%     & 28.47\%     & 67.36\%   & 44.41    & 5109  \\
    GPT4-1106    & 8.62\%    & 6.86\%   & 30.94\%     & 67.38\%   & 44.83    & 5121  \\
    InternLM-20B  & \textbf{14.89\%}   & \textbf{7.96\%}     & 31.10\%     & {\ul 67.92\%}   & \textbf{51.55}    & 5125    \\
    Qwen-14B      & 12.72\%   & 6.54\%     & 32.95\%     & 66.96\%   & 45.85    & 5125    \\
    Xinyu-7B      & 10.30\%   & 6.52\%     & 28.64\%     & 67.32\%   & 49.84    & 4978    \\
    Xinyu2-70B     & {\ul 13.41\%}   & 7.05\%     & {\ul 33.93\%}     & \textbf{68.97\%}   & {\ul 51.10}    & 5130    \\
    \bottomrule
    \end{tabularx}

    \caption{\small Discriminative, selective, and generative evaluation results. \#kws denotes the number of keywords and \#valid denotes the number of valid evaluations. In the same column, optimal values are bolded, and suboptimal values are underlined.}
    \label{tab:results}
\end{table*}

\begin{table}[]
    \small
    \centering
    \begin{tabularx}{\columnwidth}{p{0.125\textwidth}p{0.06\textwidth}p{0.06\textwidth}p{0.06\textwidth}p{0.06\textwidth}}
        \toprule
                      & KNO              & DOC                  & GEN           & NUM           \\
        \midrule
        Aquila-34B    & \textbf{59.55\%} & {\ul 54.97\%}        & 53.74\%       & 53.52\%       \\
        Baichuan2-13B & \textbf{53.75\%} & {\ul 52.10\%}        & 48.43\%       & 49.67\%       \\
        Baichuan2-53B & \textbf{57.70\%} & {\ul 57.46\%}        & 56.26\%       & 52.58\%       \\
        ChatGLM2-6B   & 40.94\%          & \textbf{45.56\%}     & {\ul 44.23\%} & 42.63\%       \\
        GPT3.5-Turbo  & \textbf{55.21\%} & {\ul 51.06\%}        & 47.63\%       & 47.85\%       \\
        GPT4-0613     & \textbf{59.87\%} & {\ul 55.99\%}        & 51.93\%       & 55.73\%       \\
        GPT4-1106     & \textbf{68.73\%} & 60.19\%              &  54.77\%      & {\ul 62.04\%} \\
        InternLM-20B  & \textbf{51.88\%} & {\ul 50.65\%}        & 49.56\%       & 48.43\%       \\
        Qwen-14B      & \textbf{62.81\%} & {\ul 57.35\%}        & 53.15\%       & 53.09\%       \\
        Xinyu-7B     & 48.44\%           & \textbf{52.02\%}     & {\ul 50.87\%} & 50.00\%       \\
        Xinyu2-70B   & \textbf{63.13\%}  & {\ul 61.47\%}        & 54.46\%       & 57.07\%       \\
        \bottomrule
    \end{tabularx}
    \caption{\small Evaluation by different types. In the same row, optimal values are bolded, and suboptimal values are underlined.}
    \label{tab:eval_by_type}
\end{table}

% 判别式测评：GPT模型的优势；关键词和判别式评测的优劣
\paragraph{Discriminative Evaluation} Initially, the GPT series models' performance is notably superior in discriminative evaluation, showcasing their formidable foundational capabilities in knowledge recall, utilization, and judgment.
Moreover, a comparison of experimental outcomes at the keyword and sentence levels reveals that accuracy is generally superior at the keyword level. This could stem from the fact that the hallucinated continuations in our dataset exhibit sufficient fluency, aligning with the fluency distribution of LLM outputs. This can potentially confuse the evaluated LLM, complicating the judgment of the continuation's authenticity. Conversely, keywords bypass fluency concerns, rendering keyword-level evaluation more amenable to LLMs. This observation implies that detecting hallucinations could be more dependable at the keyword level compared to the sentence level.

% 选择式测评：GPT和Xinyu表现不错；判别和选择的比较；跷跷板效应
\paragraph{Selective Evaluation} Firstly, GPT4-1106 clinches the top spot, reaffirming the formidable foundational capabilities of the GPT series models. Concurrently, Xinyu2-70B attains second place, excelling as a model trained on the Chinese news corpus. This achievement, to a degree, confirms the merit of domain-specific LLMs. Secondly, when comparing the outcomes of the selective evaluation with those of the discriminative evaluation at the sentence level, most LLMs exhibit improved accuracy. We think, furnishing LLMs with more contrasting information alleviates the demand for the model's fact recall, thus diminishing the challenge of selective evaluation. Therefore, we posit that selective evaluation is comparatively simpler for LLMs. Thirdly, a decline is observed in discriminative evaluation outcomes from GPT4-0613 to GPT4-1106, whereas selective evaluation outcomes register a notable increase of around 5\%. This substantiates the ``seesaw phenomenon,'' wherein certain capabilities are enhanced while others may regress, in tandem with the model's upgrade~\cite{GPTFathom_2023_arXiv_ByteDance}. This suggests that the decision to either enhance a single capability individually or to balance multiple capabilities is critical.

% 生成式测评：只分析了各个模型的表现差异
\paragraph{Generative Evaluation} Overall, InternLM-20B, Xinyu2-70B, and Aquila-34B have achieved commendable results, but the performance of Aquila-34B could be attributed to its comparatively shorter average generation length. Additionally, the GPT series exhibits subpar performance, possibly due to the insubstantial amount of Chinese data in its training corpus. After all, the Chinese data incorporated into GPT's training from the Common Crawl corpus comprises less than 5\%\footnote{\url{https://commoncrawl.github.io/cc-crawl-statistics/plots/languages.html}}.

% 按类别的评测结果
\paragraph{Evaluations by Type} We focus on selective evaluation results and perform a comprehensive breakdown analysis of these across the four types, as illustrated in Table~\ref{tab:eval_by_type}. Initially, most LLMs demonstrate enhanced accuracy for knowledge-intensive and document-intensive news. This may be because the training datasets for LLMs typically include substantial human knowledge and official documentation of major historical events. Furthermore, the majority of LLMs show reduced accuracy in general and number-intensive news. General news often contains societal minutiae, which are not the focus of LLM training. Regarding number-intensive news, it poses a considerable challenge for LLMs, given that encoding identical numbers with varied historical meanings is complex. However, GPT4-1106 attains especially high scores in the demanding number-intensive news.

\subsection{Further Discussion}

% 评测方法优劣分析
Each of the three evaluation forms possesses distinct advantages and drawbacks. Discriminative evaluation is often the method of choice for a range of standard benchmarks~\cite{HaluEval_2023_arXiv_RUC, HalluQA_2023_arXiv_Fudan}. This approach is intuitive, and the construction of evaluation prompts is straightforward. Selective evaluation resembles discriminative evaluation but is marginally less demanding because it includes a reference option for contrast. In both discriminative and selective evaluations, certain models might be suspected of conjecturing answers from a few shots due to inadequate reasoning skills, which can undermine the reliability of the outcomes. Consequently, the use of explainable prompting becomes essential. Generative evaluation most closely mirrors real-world applications. However, the generated content is unrestricted, which poses challenges for even the most dependable reference-based evaluation techniques. Therefore, employing a combination of metrics simultaneously, including lexical evaluation based on token coverage and semantic evaluation based on textual similarity, is imperative.

% 评测方法难度评价
The foundational capabilities required of LLMs can be arrayed on a spectrum from simple to complex: generative, selective, and discriminative evaluation. Generative evaluation entails the direct invocation of parameters for continuation, bypassing the need for an extensive grasp of instructions. Selective evaluation necessitates a degree of inferential reasoning but offers comparative choices, rendering the level of difficulty moderate. Conversely, discriminative evaluation demands the precise retrieval of facts, thereby increasing the challenge.

% 评测方法适用场景
Moreover, various evaluations cater to different application contexts. Should the objective be to solely improve the model's capacity for reliable continuation, generative evaluation would suffice. In the training of a dependable chatbot, selective and discriminative evaluations prove suitable. When aiming to train a reward model, selective evaluation is beneficial, offering evaluation for positive and negative instances. If the goal is to enhance the model's ability to recall and apply knowledge, discriminative evaluation emerges as the demanding option.

\section{Conclusion}

% 大模型厉害，需要benchmark，我们的benchmark好，做了实验，未来工作
LLMs are rapidly evolving, heralding a new era of potential applications within the realm of professional content generation. The progression of LLMs in this domain necessitates the establishment of robust benchmarks to steer their development effectively. In this work, we introduce a novel hallucination benchmark dataset using an unconstrained fashion, encompassing more than 5,000 instances annotated at the keyword level. Additionally, we propose a secure, scalable, and user-friendly evaluation framework to facilitate comprehensive assessments. Through meticulous experimentation on eleven prominent LLMs, our study has unearthed a series of enlightening findings. Looking ahead, our research endeavors will persist in exploring the intricacies of hallucination phenomena within professional content generation, aiming to further understand and enhance LLM capabilities.

\section*{Limitations}

\paragraph{Dataset} Firstly, although we have utilized hallucination ranking, automatic labeling, human rechecking, and various other techniques mentioned in Appendix~\ref{apdx:uhgeval_dataset} to ensure the quality of data annotation, with over 5,000 data entries, there is still a possibility of labeling errors. We have mobilized the power of the open-source community to collectively improve our dataset. Secondly, the dataset creation process is flexible, allowing for dataset expansion into English and broader domains, such as mathematical reasoning and programming codes. Thirdly, a minor error in the dataset creation process has resulted in a relatively unbalanced distribution of the dataset across the five different LLMs used for generation. A detailed analysis of this issue can be found in Appendix~\ref{apdx:imbalance}.

\paragraph{Framework} Although our framework simplifies the integration of LLMs through APIs or vLLM\footnote{\url{https://github.com/vllm-project/vllm}}, users seeking to utilize custom or diverse HuggingFace models may face initial hurdles. We need to further enhance the usability of our framework.

\paragraph{Constrained v.s. Unconstrained} We have determined that constrained generation cannot fully reflect real-world applications, but empirical analysis is required to prove this point. This may involve constructing a text classifier to determine the type of hallucination, followed by comparing the distribution of hallucinations in our dataset with those in other benchmark datasets to observe any significant deviations. We leave this for future work.

\section*{Acknowledgments}

This work was supported by the National Natural Science Foundation of China (Grants No. 62072463, 71531012), the National Social Science Foundation of China (Grants No. 18ZDA309), the Research Seed Funds of the School of Interdisciplinary Studies at Renmin University of China, and the Opening Project of the State Key Laboratory of Digital Publishing Technology of the Founder Group.

\bibliography{references}

\newpage
\onecolumn
\appendix
\appendixpage
\startcontents[sections]
\printcontents[sections]{l}{1}{\setcounter{tocdepth}{2}}
\twocolumn

\clearpage
\section{Comparisons with Other Datasets} \label{apdx:comparsions_datasets}

Below are specific comparisons with other datasets and the significance of unconstrained generation.

\subsection{TruthfulQA}

The TruthfulQA dataset encompasses three modes of evaluation, with the primary mode being generative. In this mode, a problem is presented to the model, which then freely generates content that is assessed by humans or a fine-tuned GPT-judge. The other two modes are single- / multiple-choice questions. In these modes, a problem along with reference options is provided, the model makes a selection, and accuracy is calculated.

Figure 1 in the TruthfulQA paper includes statements indicating that some content is freely generated by GPT-3. This might be somewhat misleading. The content is used solely to evaluate the performance of the GPT-3 model in generative evaluation and is not part of the dataset. The actual free generation pertains to the "reference options" in the single- / multiple-choice questions. These reference options are manually crafted in TruthfulQA.

Appendix C of the paper details the method used to create the reference options:

\begin{quote}
\textit{Reference answers for each question in TruthfulQA are constructed as follows:}

\textit{We take a set of true answers directly from Wikipedia (or the listed source). We then try to provide coverage of common variations on this answer...}

\textit{We follow a similar process for generating false answers, but widen the answer set by running internet searches for [common misconceptions / superstitions / conspiracies around X] where relevant, as there tend to be many possible imitative false answers that are not always covered in a single source...}
\end{quote}

\subsection{HaluEval}

The problem types within this benchmark are all judgment questions, tasked with determining whether an option contains hallucinations. Accordingly, they also provide reference options. However, their method of generating these options is targeted. An example of how they generate options is: "You are trying to answer a question but misunderstand the question context and intention." They then take such generated texts and real texts, placing them together for downstream models to evaluate for the presence of hallucinations.

\subsection{HaDes}

HaDes evaluates a model's ability to identify hallucinated words within a given text. However, the method used to generate these hallucinations involves randomly altering correct text, thereby transforming some words into hallucinations. This approach to generating errors leads to a distributional bias compared to the hallucinations that arise from the model's free output.

In summary, most existing datasets related to hallucinations are purposefully and manually generated with constraints. They do not represent the hallucinations that might be collected while the model is addressing user queries or responding to users in real-world scenarios. This raises the question: Are the errors generated in this manner truly reflective of the mistakes a model would make? Hence, in creating our dataset, we allowed the model to output freely, collecting only those portions where hallucinations occurred. This represents one of the major challenges in our work.

\subsection{Why Is Unconstrained Generation Important?}

In datasets like TruthfulQA, HaluEval, and HaDes, it's challenging to pinpoint exactly why a model might produce hallucinations. These texts, potentially containing inaccuracies, are designed to assess whether a downstream model can identify errors within a text. However, our dataset enables a genuine evaluation of model hallucinations, even tracing their origins. For instance, in our dataset, the entry with ID doc\_000002 features hallucinations generated by the Baichuan2-13B model. The terms related to "economic development" and others, totaling five words, are involved in these hallucinations, while words like "China" and another set of five words are not. This distinction allows us to investigate whether there are differences in the token logits, the states of hidden layers, etc., between the words associated with hallucinations and those without, in the context of the Baichuan model. Theoretically analyzing the causes of hallucinations within the Baichuan model is part of our ongoing work. This approach is something that other benchmarks cannot offer, as their hallucinations are not freely produced by the model, and in some cases, not even generated by models.

\clearpage
\section{More Related Works} \label{apdx:related}

\subsection{Large Language Models}

% LLM的能力
Language models are pivotal in computer science, evolving from statistical language models to neural language models, to pre-trained language models (PLMs), and now to the current generation of LLMs. The advent of models such as ChatGPT has seen contemporary LLMs exhibit new capabilities in handling complex tasks. These models can manage few-shot tasks via in-context learning and tackle mixed tasks by following instructions~\cite{LLMSurvey_23_arXiv_RUC}.

% LLM的分类，我们用了哪些
LLMs can be classified according to two dimensions. The first dimension concerns the openness of the model weights. For example, open-source models include Meta's LLaMA~\cite{LLaMA2_2023_arXiv_Meta}, Tsinghua University's GLM~\cite{GLM_22_ACL_CCFA}, and Alibaba's Qwen~\cite{Qwen_23_arXiv_Alibaba}, while closed-source models feature OpenAI's GPT~\cite{GPT4_23_arXiv_OpenAI}, Baidu's ERNIE Bot~\cite{ERNIE3_2021_arXiv_Baidu}, and Anthropic's Claude\footnote{\url{https://www.anthropic.com/index/introducing-claude}}, among others. The second dimension differentiates between the use of a PLM or a supervised fine-tuned (SFT) model for specific inferences~\cite{zhu2024proxy}. A PLM is a language model trained on extensive unlabeled textual data to discern underlying patterns, structures, and semantic knowledge within the corpus. Conversely, an SFT model involves further training a PLM with labeled datasets tailored to a specific task, to improve performance in that area. Many open-source models, including LLaMA, GLM, and Qwen, have made their PLM weights publicly available. For SFT models, users can access the chat variants of open-source models or the API services provided by closed-source models. In our research, we focus primarily on evaluating closed-source GPT series models and open-source Chinese SFT models.

\subsection{Hallucinations in LLM}

% 幻觉的概念，分类
Despite remarkable advancements in LLMs, they continue to encounter challenges, with hallucination being one of the most notable. Hallucination in language models refers to generating content that strays from factual accuracy, leading to unreliable outputs. Hallucinations occur when the generated content is not aligned with user input, deviates from the model's previous outputs, or is at odds with established real-world knowledge~\cite{HalluSurvey_23_arXiv_Tencent}. 

% 幻觉的表现
Specific examples include inaccuracies in age, currency, scores, and other numerical values; citing fictional statements; inventing non-existent characters; and muddling timelines by merging events from different periods~\cite{HalluSurvery2_2023_arXiv_SouthCarolina}. 

% 幻觉的原因
Regarding the causes of hallucinations, several factors can be responsible~\cite{HalluSurvey_23_arXiv_Tencent}. One contributing factor is the use of inaccurate or incomplete training data. During training, LLMs fine-tune their parameters with vast quantities of text data. However, this data may be flawed, harboring errors, inaccuracies, or gaps in information. Another factor involves inconsistencies in contextual information. While LLMs typically consider previously generated context when producing content, challenges in managing long-term dependencies or understanding complex contexts can result in inconsistencies. Additionally, hallucinations can arise from lacking or erroneous world knowledge. Although LLMs gain considerable world knowledge via training data, they may be deficient in specific domain knowledge or misinterpret certain facts, leading to hallucinations. Furthermore, model limitations, including generation strategies and alignment methods, can also play a role in hallucinations during content creation.

\clearpage
\section{The UHGEval Dataset} \label{apdx:uhgeval_dataset}

\subsection{Dive into Human Rechecking Process} \label{apdx:human}

% 语言依赖结构和最小幻觉原则
\paragraph{Least Hallucination Principle} The keyword-based labeling scheme has inherent limitations. Languages exhibit a dependency structure~\cite{DependencyGrammar_2019_AnnualReviewOfLinguistics_OhioU}. For instance, in the phrase ``The rainbow is black,'' the words ``rainbow'' and ``black'' exhibit interdependence. One could contend that ``black'' is incorrect, while another could maintain that ``rainbow'' is erroneous, given that ``night'' is typically described as black. To address the challenges stemming from language dependency structures, we have adopted the \textit{Least Hallucination Principle}. If a set of words can be selected, and their replacement with contextually appropriate words yields a reasonable sentence, then such a set of words is designated as a hallucinated word group. The words selected for annotation must meet the condition of comprising the minimal number of words in the group, as illustrated in Equation~\ref{eq:least_hallu}.  In the equation, $\mathbf{W}$ is the set of keywords in a sentence, $\mathbf{w}$ is the hallucinated word group, $\text{correct}(\cdot)$ is the correction function that modifies hallucinated words to non-hallucinated words, and $\text{hallucinated}(\cdot)$ assesses whether a sentence composed of keywords hallucinated.

\begin{equation} 
    \begin{split}
        \min\quad & |\mathbf{w}| \\
        \mbox{s.t.}\quad & \mathbf{w} \subset \mathbf{W} \\
        & \mathbf{w}' = \text{correct}(\mathbf{w})  \\
        & \text{false} = \text{hallucinated}(\mathbf{W}-\mathbf{w}+\mathbf{w}') \\
    \end{split}
    \label{eq:least_hallu}
\end{equation}

% 举例  
\noindent By this principle, within the phrase ``Journey to the West is an American novel and one of the Four Great Classics,'' the word ``American'' would be marked for annotation, as altering this single keyword to ``Chinese'' dispels the hallucination throughout the sentence.

% 标注员的参与感
\paragraph{Engagement of Annotators} Additionally, we acknowledge that hallucination annotation may become somewhat tedious. Consequently, annotators are integrated throughout the entire process, participating in discussions instead of solely evaluating the accuracy of machine annotations. This approach also yields benefits for our work. For example, an annotator with a journalism background offered valuable professional insights into pinpointing news-related hallucinations, emphasizing that fact increment is a critical aspect of news writing.

% 标注人员组成，价格
\paragraph{Annotation Team} Our annotators are all Chinese nationals with Chinese as their native language, each holding at least a Master's degree in Journalism. We collaborated with a well-known, sizable news organization in China, Xinhua News Agency. Some of their staff joined our research team and participated in data annotation for this project. There were a total of 9 annotators involved in this project, with a gender ratio of 1:2 (male to female). Regarding their compensation, they first received a standard employee salary. Additionally, they were paid an extra 3 RMB for each data item annotated, with each item taking about 40 seconds to annotate. Besides the annotators, our engineering team and experts from the journalism industry at Xinhua News Agency participated in the data review process, totaling 3 people. Our main responsibility was to supervise and review the quality of the annotations. The entire annotation process lasted for 22 days.

\subsection{Analysis of the Final Dataset} \label{apdx:data}

\begin{figure}[h]
    \centering
    \includegraphics[width=0.97\linewidth]{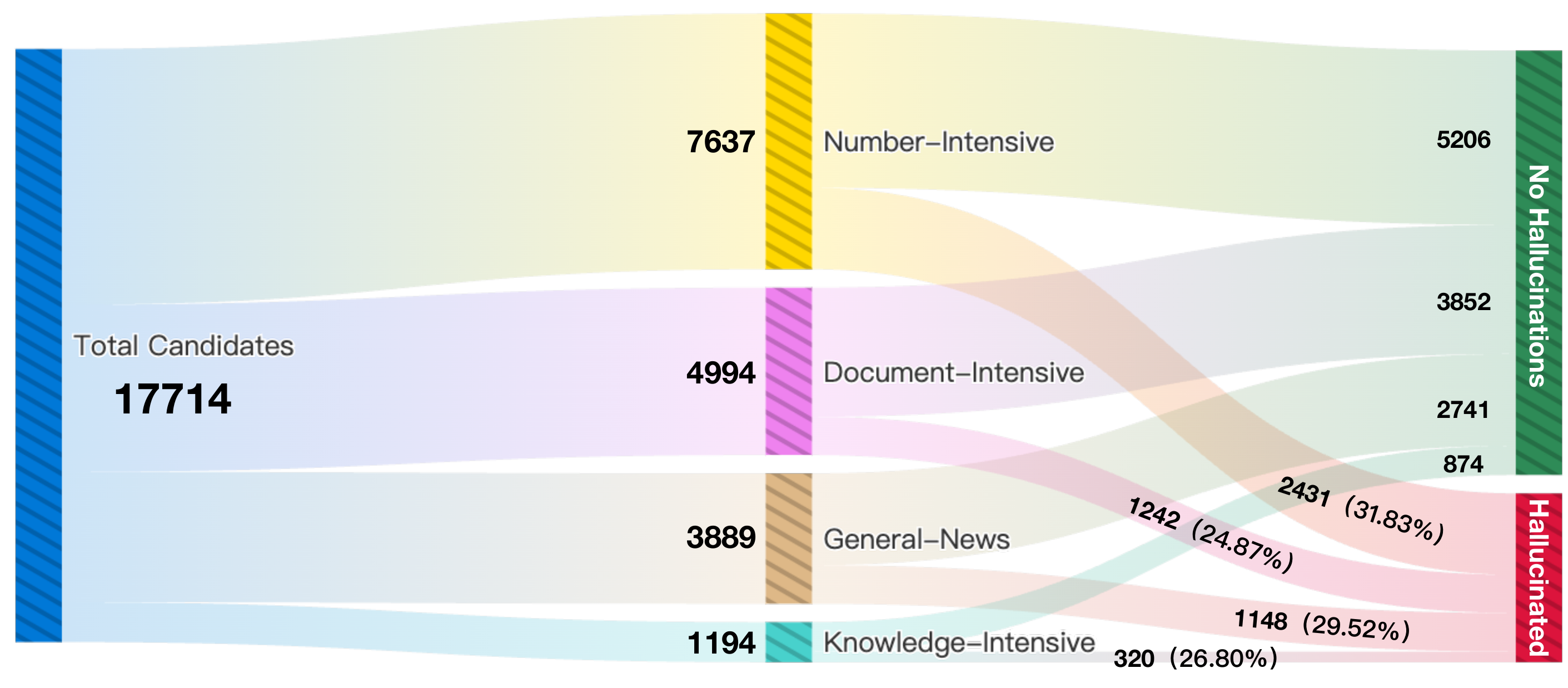}
    \caption{\small Conversion rates from candidates to hallucinations.}
    \label{fig:conv_rate}
\end{figure}

% 转化率
We developed a conversion rate chart to depict the transition from candidate hallucinations to the final dataset, as depicted in Fig.~\ref{fig:conv_rate}. The conversion rate can be interpreted as the likelihood of hallucinations occurring across various categories. Our observations indicate a higher likelihood of hallucinations in number-intensive and general news, whereas this likelihood is reduced in knowledge-intensive and document-intensive news.

\vspace{-2mm}

\begin{figure}[h]
    \centering
    \includegraphics[width=0.97\linewidth]{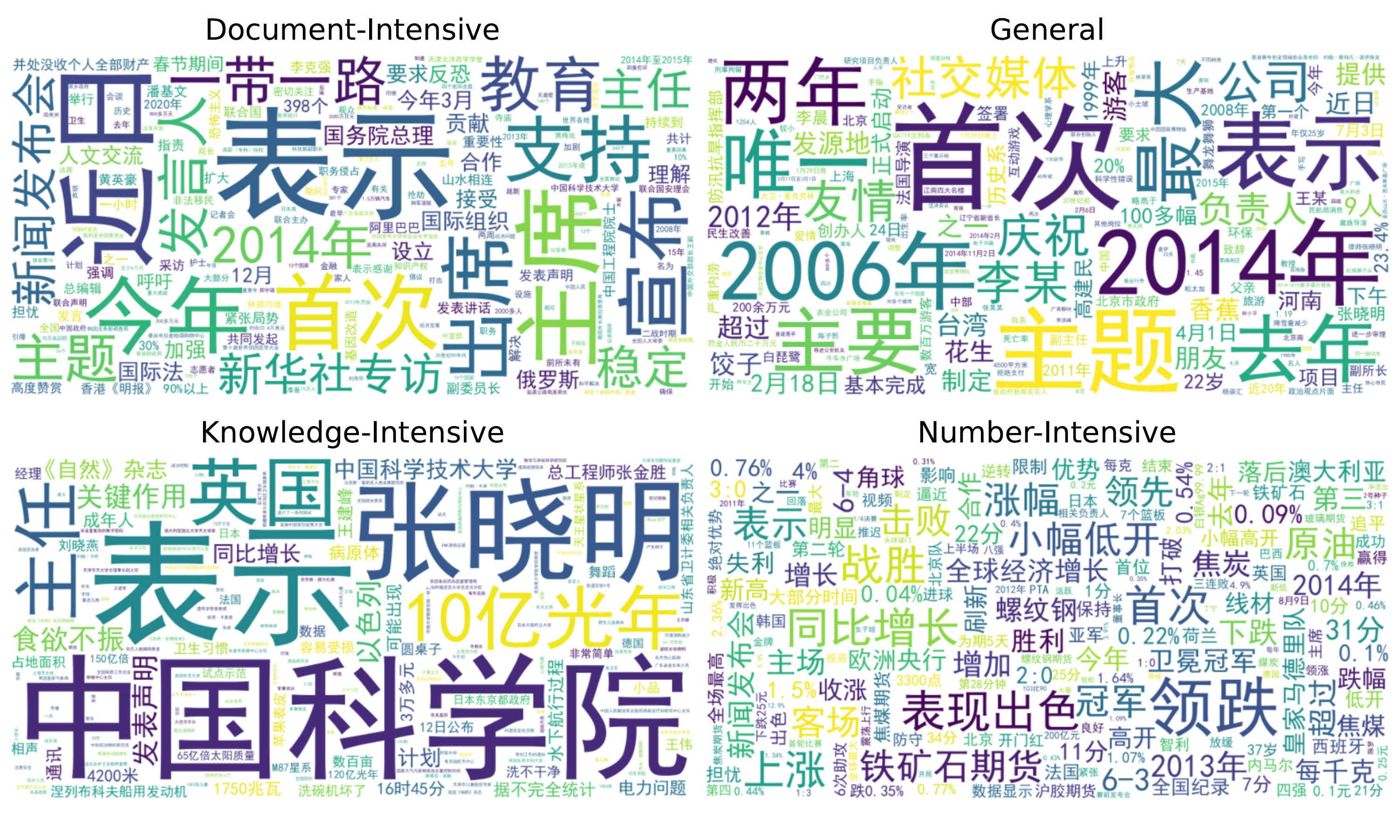}
    \caption{\small Hallucinated keywords in different types of news}
    \label{fig:word_cloud}
\end{figure}

% 词云分析
By analyzing the hallucinated word cloud depicted in Fig.~\ref{fig:word_cloud} for each news category, we can draw the following conclusions: Number-intensive news often includes numeric values that are challenging to remember, like 0.09\% and 6:3, which pose difficulties for both LLMs and humans. General news encompasses a diverse vocabulary, featuring terms such as ``social media'' and ``friendship,'' which are often deemed less critical and thus challenging to incorporate into the training corpora of many LLMs. Knowledge-intensive news frequently features terms such as ``according to incomplete statistics'' and ``key technology,'' which are prevalent in technical literature. However, LLMs may not always use these terms appropriately. Document-intensive news often contains terms associated with official statements, such as ``representation,'' ``president,'' and ``spokesperson.'' This suggests that LLMs are susceptible to introducing unauthorized alterations to the content of documents.

\subsection{Data Volume for Each Step} \label{apdx:data_volume}

In this section, we present the data volume at various stages of our dataset creation process for reference and transparency.

\begin{itemize}
\item Data volume of original news dataset: 737,766
\item Data volume after preprocessing: 25,005 (filtering out outliers in dimensions such as length and news type)
\item Data volume after generating candidate hallucination text: 17,503 (filtering out data items that did not generate appropriate continuations, such as those with excessively short length or insufficient extracted keywords)
\item Data volume with hallucination in machine-labeled text: 8,314 (filtering out texts deemed by the machine to lack hallucination)
\item Data volume after human annotator labeling: 5,141 (filtering out instances not verified as hallucination upon manual review, or deemed inappropriate, such as those repeating previous content or generating text no longer of news type but rather comprehension questions)
\end{itemize}

\onecolumn
\subsection{An example from the UHGEval Dataset} \label{apdx:an_example}

\begin{figure}[h]
    \centering
    \includegraphics[width=\linewidth]{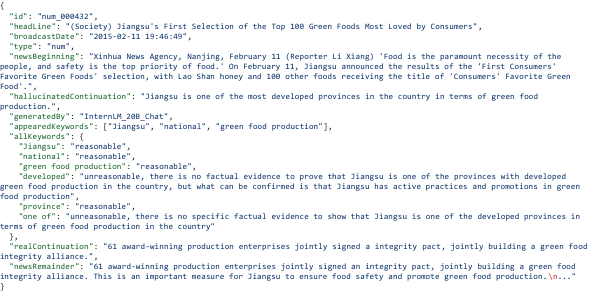}
    \caption{An example from the UHGEval dataset. (In English)}
    \label{fig:dataset_example}
\end{figure}

\begin{figure}[h]
    \centering
    \includegraphics[width=\linewidth]{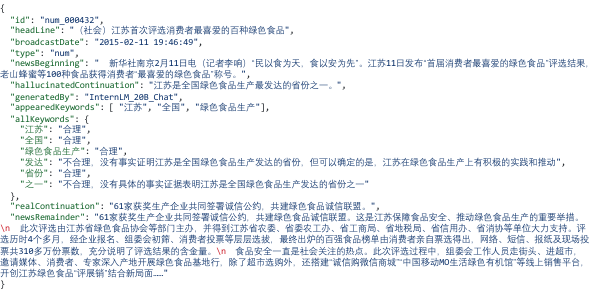}
    \caption{An example from the UHGEval dataset. (In Chinese)}
    \label{fig:dataset_example_ch}
\end{figure}

\twocolumn

\clearpage
\section{Experiments}

\subsection{LLMs Employed in This Research} \label{apdx:models}

All LLMs used in this study are detailed in Table~\ref{tab:models}.

\begin{table}[h]
    \small
    \centering
    \begin{tabular}{lllll}
        \toprule
        Model           & \#Para.           & Publisher     & Date  \\ 
        \midrule
        GPT3.5-Turbo    & 175B$^*$          & OpenAI        & 2023.03$^*$ \\
        GPT4-0613       & NaN               & OpenAI        & 2023.06 \\
        ChatGLM2        & 6B                & Tsinghua      & 2023.06\\ 
        Xinyu           & 7B                & IAAR\&Xinhua  & 2023.06 \\ 
        InternLM        & 20B               & ShLab         & 2023.07 \\ 
        Baichuan2       & 13B               & Baichuan Inc. & 2023.09 \\ 
        Baichuan2       & 53B               & Baichuan Inc. & 2023.09 \\ 
        Qwen            & 14B               & Alibaba       & 2023.09 \\ 
        Aquila2         & 34B               & BAAI          & 2023.10 \\
        Xinyu2          & 70B               & IAAR\&Xinhua  & 2023.10 \\
        GPT4-1106       & NaN               & OpenAI        & 2023.11 \\
        \bottomrule
    \end{tabular}
    \caption{LLMs sorted by release date. All LLMs are chat models. Asterisk (*) denotes estimated value, NaN denotes no public data available, and 175B denotes 175billion.}
    \label{tab:models}
\end{table}

% 基座介绍
GPT represents a series of LLMs developed by OpenAI~\cite{GPT4_23_arXiv_OpenAI}. In this study, GPT3.5-Turbo, GPT4-0613, and GPT4-1106 are utilized. GLM constitutes a pre-training framework proposed by Tsinghua University~\cite{GLM_22_ACL_CCFA}, and the ChatGLM2-6B chat model is employed. InternLM serves as an open-source, lightweight training framework, with its development team releasing a spectrum of models utilizing this framework~\cite{InternLM_23_GitHub_InternLMTeam}; the InternLM-20B open-source chat model is utilized in the present work. Baichuan2 comprises a series of expansive, multilingual base language models~\cite{Baichuan2_23_arXiv_Baichuan}, with both the open-source Baichuan2-7B chat model and the closed-source Baichuan2-53B chat model being employed in this investigation. Qwen encompasses a language model series characterized by distinct models with varying parameter counts~\cite{Qwen_23_arXiv_Alibaba}, and the Qwen-14B open-source chat model is utilized in the current study. Aquila2 represents a language model series devised by BAAI, noted for surpassing comparable models in terms of performance~\cite{Aquila2_23_GitHub_BAAI}, and the Aquila2-34B chat model is employed in this research. 

Besides, the Xinyu series models are the results of a collaborative research and development effort between the Institute for Advanced Algorithms Research, Shanghai (IAAR, SH), and the State Key Laboratory of Media Convergence Production Technology and Systems of the Xinhua News Agency. Xinyu-7B is an augmented large-scale language model derived from the foundational model, BloomZ-7B~\cite{BLOOMZ_2023_arXiv_HF} through continued pre-training, news-specific fine-tuning, and alignment optimization. And, Xinyu2-70B is developed based on the open-source LLaMA2-70B~\cite{LLaMA2_2023_arXiv_Meta} framework, incorporating expansions to the Chinese lexicon, ongoing pre-training, and news-specific fine-tuning, thereby endowing it with a robust foundational capability in the news domain.

\subsection{Evaluation Method} \label{apdx:eval}

% 幻觉评测的三个角度: form, metric, granularity
The evaluation of hallucinations can be decomposed into three principal dimensions: form, metric, and granularity. Form concerns how the model interacts with the evaluation dataset; metric refers to the precise computational approach utilized for performance assessment; and granularity signifies the depth of detail considered in the evaluation of hallucinations. 

% form 角度 + human evaluation
\paragraph{Form} This encompasses human evaluation, discriminative evaluation, selective evaluation, and generative evaluation, among others. Human evaluation entails the direct application of human judgment to determine if the model's output contains hallucinations, representing a critical evaluation form~\cite{LLMEvalSurvery_2023_arXiv_JilinU}. However, the drawbacks of this approach are evident: evaluating too many data points is tantamount to annotating a new dataset, with the associated time and financial expenditures proving prohibitive.

% discriminative evaluation
Discriminative evaluation enables LLMs to respond with binary answers of ``yes'' or ``no''~\cite{HaluEval_2023_arXiv_RUC, HalluQA_2023_arXiv_Fudan}. Specifically, this evaluation modality involves presenting the LLM under scrutiny with an initial text followed by a continuation that may or may not include hallucinations. The LLM is tasked with producing a verdict as to the presence of hallucinations. Owing to the efficacy of few-shot prompting, this evaluation paradigm is relatively uncomplicated for LLMs to administer, as it facilitates the elicitation of the requisite responses. However, this method depends solely on the LLM's ability to draw upon the knowledge encoded within its parameters, necessitating the concurrent application of knowledge and reasoning, and thus requiring a robust foundational model capacity.

% selective evaluation
Selective evaluation allows LLMs to tackle multiple-choice questions by choosing between option A or B, as exemplified by PandaLM~\cite{PandaLM_2023_arXiv_PKU}. Specifically, in selective evaluation, the LLM under evaluation is presented with an initial text followed by two continuations: one that includes hallucinations and another that does not. The LLM's objective is to identify which of the two is hallucinated. This assessment method offers the LLM more contextual information than discriminative evaluation, thereby alleviating the burden of fact-checking and lessening the dependence on retrieving facts from its parameters. Consequently, this reduces the level of difficulty for the LLM.

% generative evaluation
However, both discriminative and selective evaluations encounter a substantial challenge. They are predicated on the assumption that ``LLMs's capacity to produce reliable text is contingent upon their discernment between hallucinated and non-hallucinated content.'' These methods do not simulate the evaluation of the model's output for hallucinations. Consequently, generative evaluation is crucial as it directly evaluates the presence of hallucinations in the text generated by the LLM under evaluation. However, the challenge arises from the fact that it is not feasible to automatically and accurately ascertain if the newly generated text is hallucinated; if it were, annotated datasets would be redundant. In scenarios of unrestrained text generation, this issue becomes increasingly complex. This complexity stems from the fact that text generated without constraints may introduce a multitude of entities and facts absent in the reference material, complicating the verification of their accuracy. Despite these hurdles, generative evaluation continues to be a predominant strategy in Natural Language Generation (NLG) tasks~\cite{NLGEval_2017_ACL_HeriotWattU}.

% metric 角度
\paragraph{Metric} Metrics include classification metrics such as accuracy, precision, recall, and others, which are applicable to human evaluation, discriminative evaluation, and selective evaluation. Generative evaluation, on the other hand, encompasses both lexical and semantic metrics. Lexical metrics evaluate the extent of token overlap between the generated text and the reference information, including metrics such as BLEU~\cite{BLEU_2002_ACL_WatsonResearch}, ROUGE~\cite{ROUGE_2004_ACL_USouthernCalifornia}, and the newly proposed metric by us, kwPrec. Semantic metrics gauge the similarity in meaning between sentences, with examples including BERTScore~\cite{BERTScore_2020_ICLR_CornellU}, GPT-judge~\cite{TruthfulQA_2022_ACL_Oxford}, and GPTScore~\cite{GPTScore_2023_arXiv_NUS}, among others.

% 粒度角度
\paragraph{Granularity} Evaluations can be conducted at both the sentence and keyword levels. Owing to our annotation methodology, our dataset is marked at the keyword level to signify instances of hallucinations. This approach affords a broader spectrum of possibilities for configuring the evaluation task, enabling the evaluated model to address the presence of hallucinations at either the keyword level, the sentence level, or even the document level.

\subsection{UHGEval Framework in Detail} \label{apdx:framework}

% 分层介绍
The framework comprises four ascending layers: the dependency layer, the evaluator layer, the core layer, and the interface layer. 

\textbf{The dependency layer} defines the essential foundational components needed for the evaluation framework, including datasets, LLM hubs, and various metrics. Importantly, each component is designed for extensibility: datasets can be replaced with custom ones, LLMs can be integrated via APIs or platforms like Hugging Face\footnote{\url{https://huggingface.co/models}}, and metrics can be customized to fit specific needs.

\textbf{The evaluator layer}, constituting the second layer, centers on an abstract class, Evaluator, and its various implementations. Within this layer, three distinct types are implemented: GenerativeEvaluator, DiscriminativeEvaluator, and SelectiveEvaluator. Users may also engineer custom evaluators, contingent upon adherence to the interface specifications of the abstract class, necessitating merely three function overloads. 

\textbf{The core layer}, representing the third stratum, comprises two principal modules: experiment.py and analyst.py. The former facilitates experiments involving multiple LLMs, evaluators, and processes, whereas the latter is tasked with the statistical analysis of experimental outcomes. 

\textbf{The interface layer}, serving as the final layer, orchestrates the user's interaction with $\mathbb{UHG}$Eval. To streamline the initiation process, a succinct 20-line demonstration is offered, alongside a run.py script for launching experiments through the command line.

\clearpage
\onecolumn % 防止图片到下一页
\section{Figures in Chinese}

\begin{figure}[h!]
    \centering
    \includegraphics[width=0.5\linewidth]{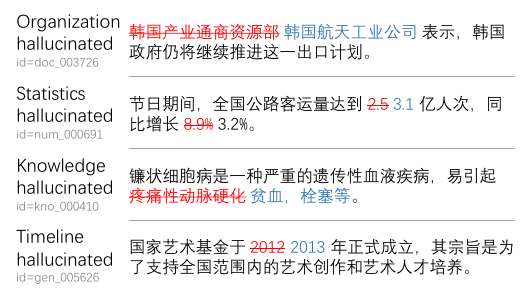}
    \caption{\small
        Hallucinations from $\mathbb{UHG}$Eval. Using the IDs, you can locate the original news articles. 
        (In English: Fig.~\ref{fig:halu_example})
    }
    \label{fig:halu_example_ch}
\end{figure}

\begin{figure}[h!]
    \centering
    \includegraphics[width=\linewidth, keepaspectratio]{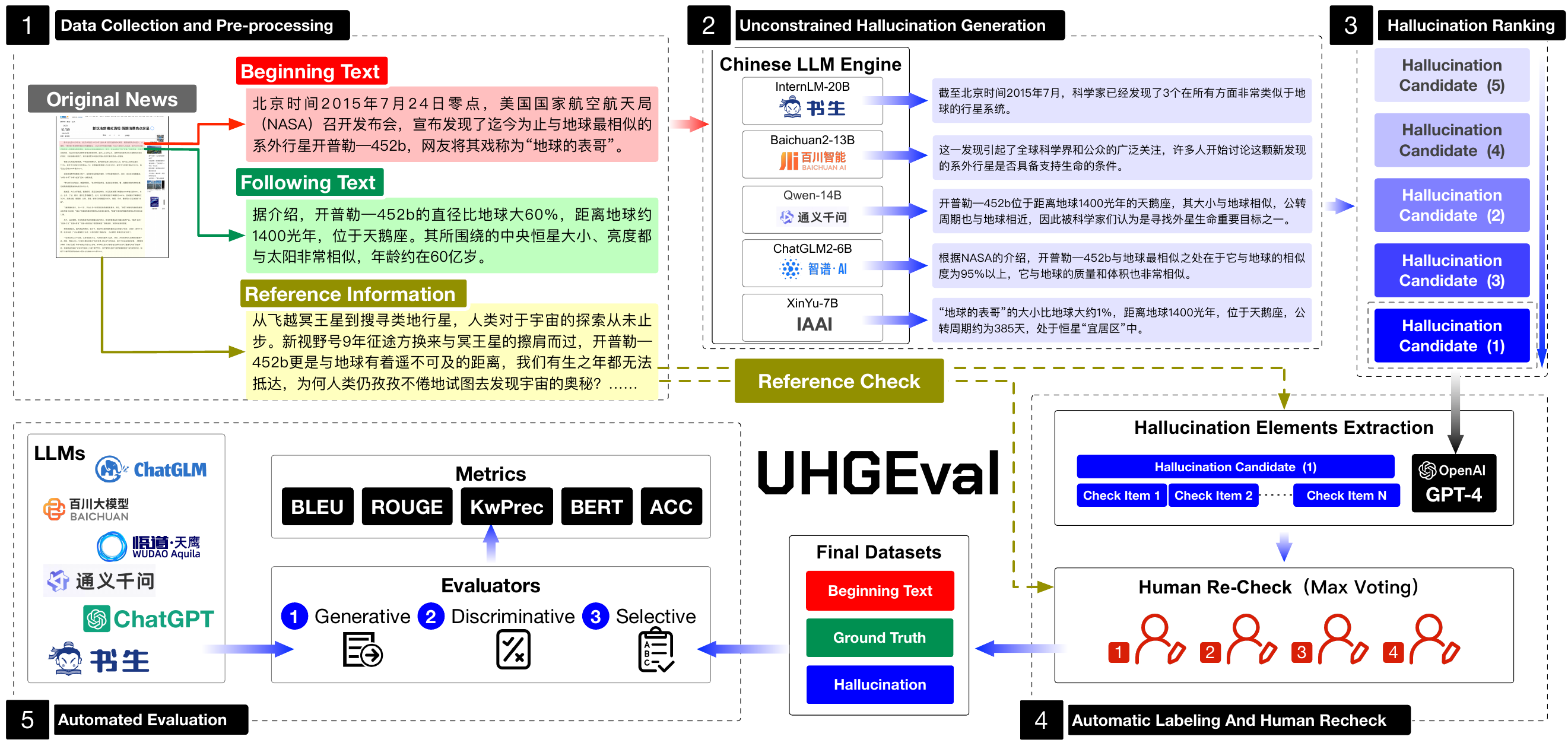}
    \caption{
        \small
        The process of creating $\mathbb{UHG}$Eval. Steps 1 to 4 regarding the creation of the benchmark dataset are explained in Section~\ref{sec:uhgeval}; Step 5, concerning the evaluation framework, is detailed in Section~\ref{sec:experiments}.
        (In English: Fig.~\ref{fig:framework})
    }
    \label{fig:framework_ch}
\end{figure}

\begin{figure}[h!]
    \centering
    \includegraphics[width=0.4\linewidth]{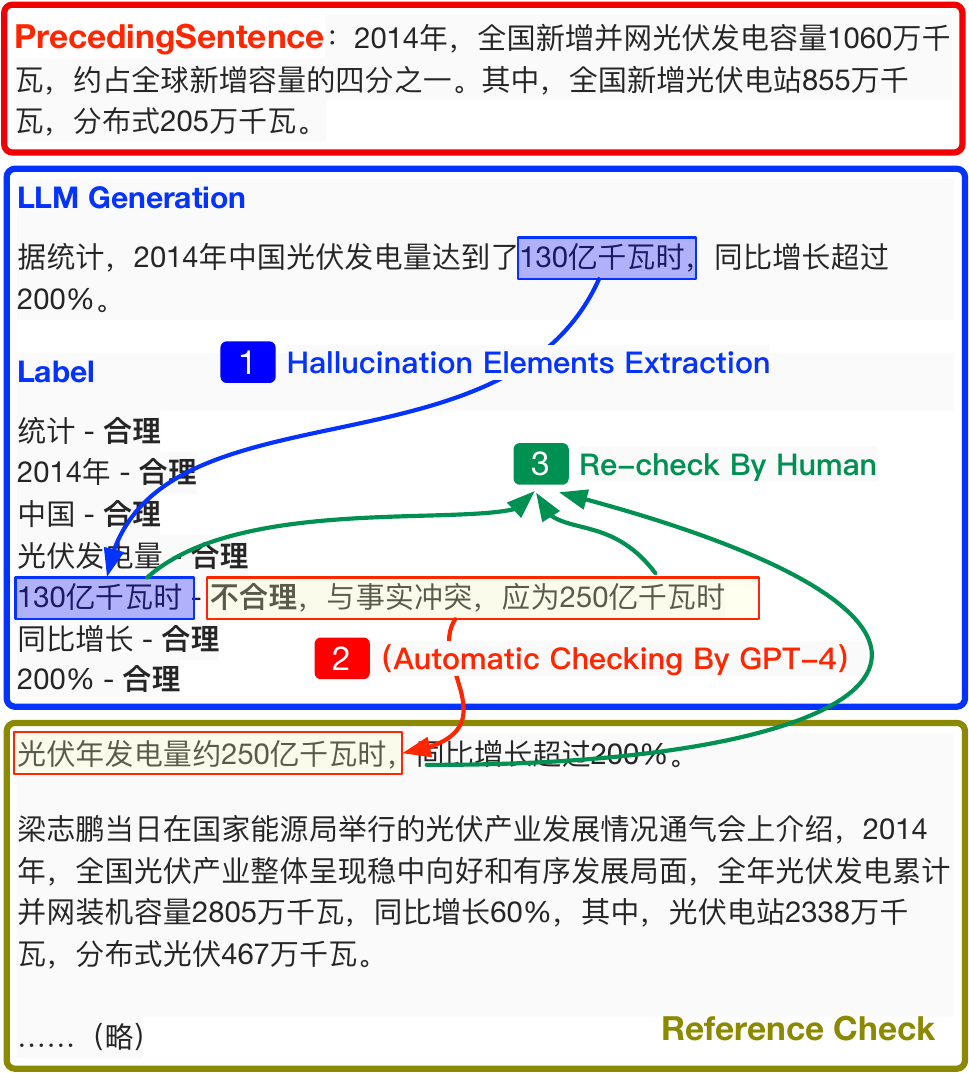}
    \caption{
        \small 
         Labeling and rechecking.
        (In English: Fig.~\ref{fig:annotation})
    }
    \label{fig:annotation_ch}
\end{figure}

\clearpage
\section{Prompt Templates} \label{apdx:tempalte}

In these templates, the orange text represents intent and instruction, the green text represents demonstrations, and the black text represents specific questions. The template may be very long, and we may use ellipses to omit some content in the middle. The original templates are in Chinese, and we also provide English translations.

\begin{figure}[h!]
    \centering
    \includegraphics[width=\linewidth]{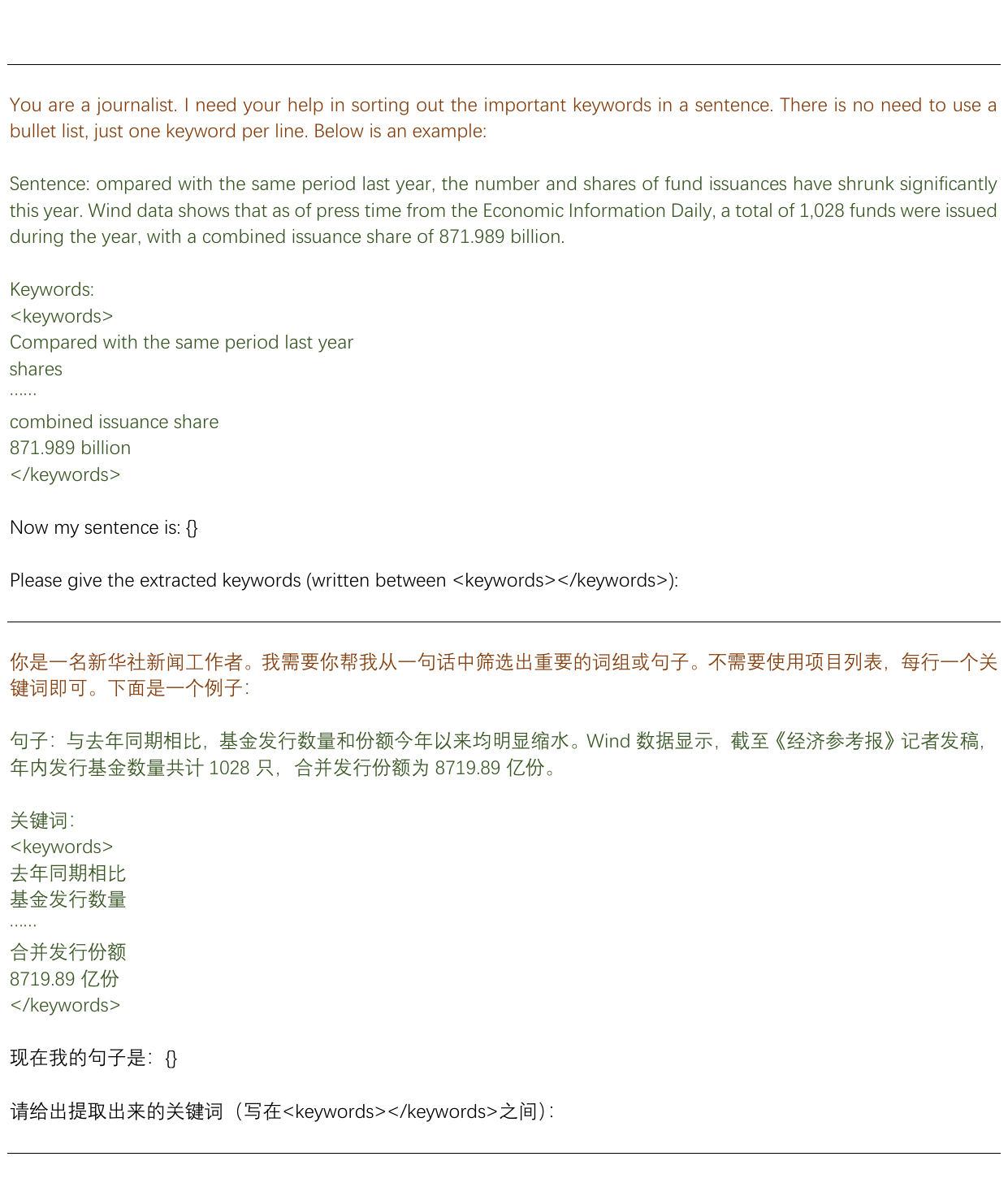}
    \caption{Prompt template for extracting keywords}
    \label{fig:templates0}
\end{figure}

\begin{figure}[h!]
    \centering
    \includegraphics[width=\linewidth]{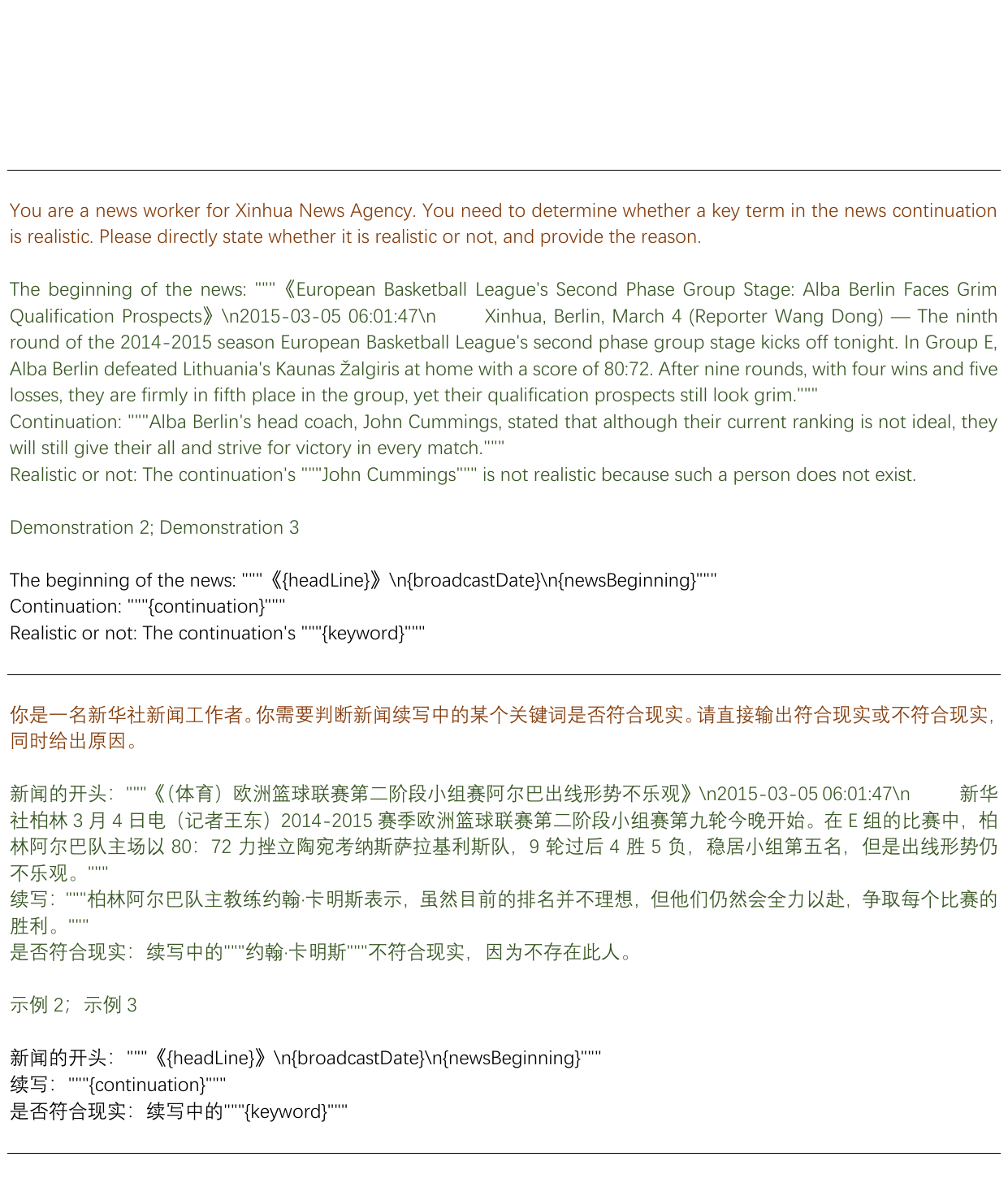}
    \caption{Prompt template for discriminative evaluation (keyword level)}
\end{figure}

\begin{figure}[h!]
    \centering
    \includegraphics[width=\linewidth]{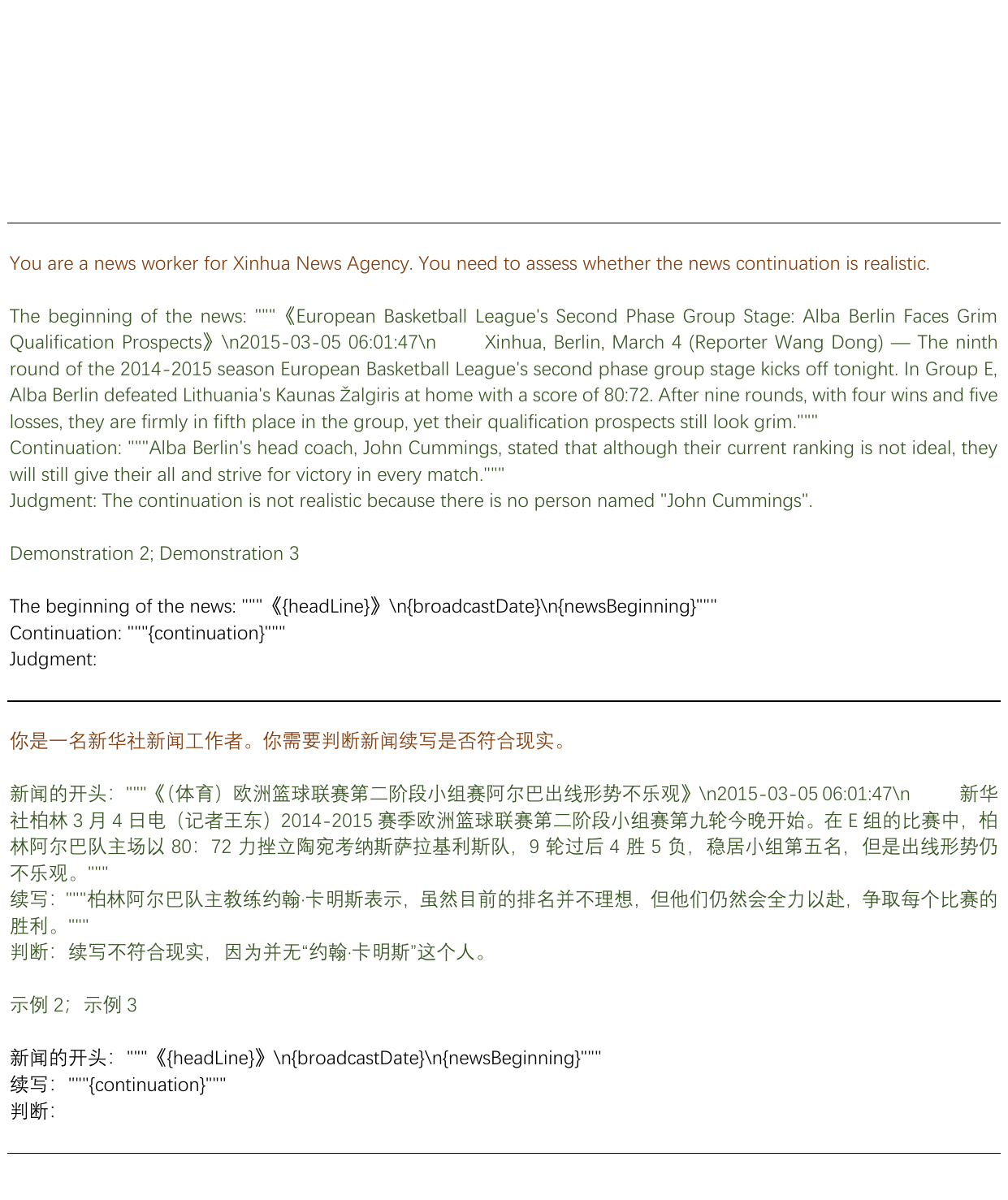}
    \caption{Prompt template for discriminative evaluation (sentence level)}
\end{figure}

\begin{figure}[h!]
    \centering
    \includegraphics[width=\linewidth]{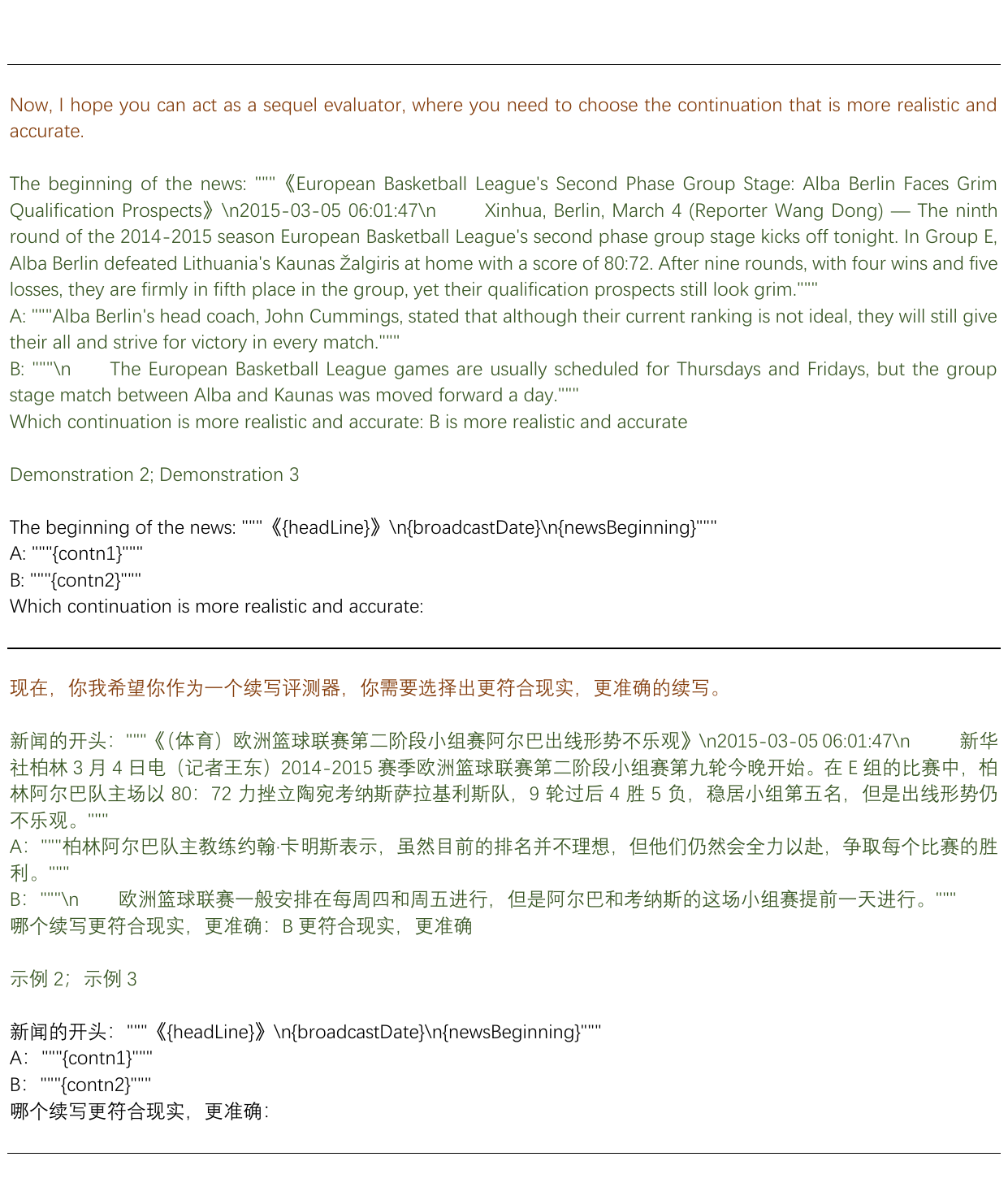}
    \caption{Prompt template for selective evaluation}
\end{figure}

\begin{figure}[h!]
    \centering
    \includegraphics[width=\linewidth]{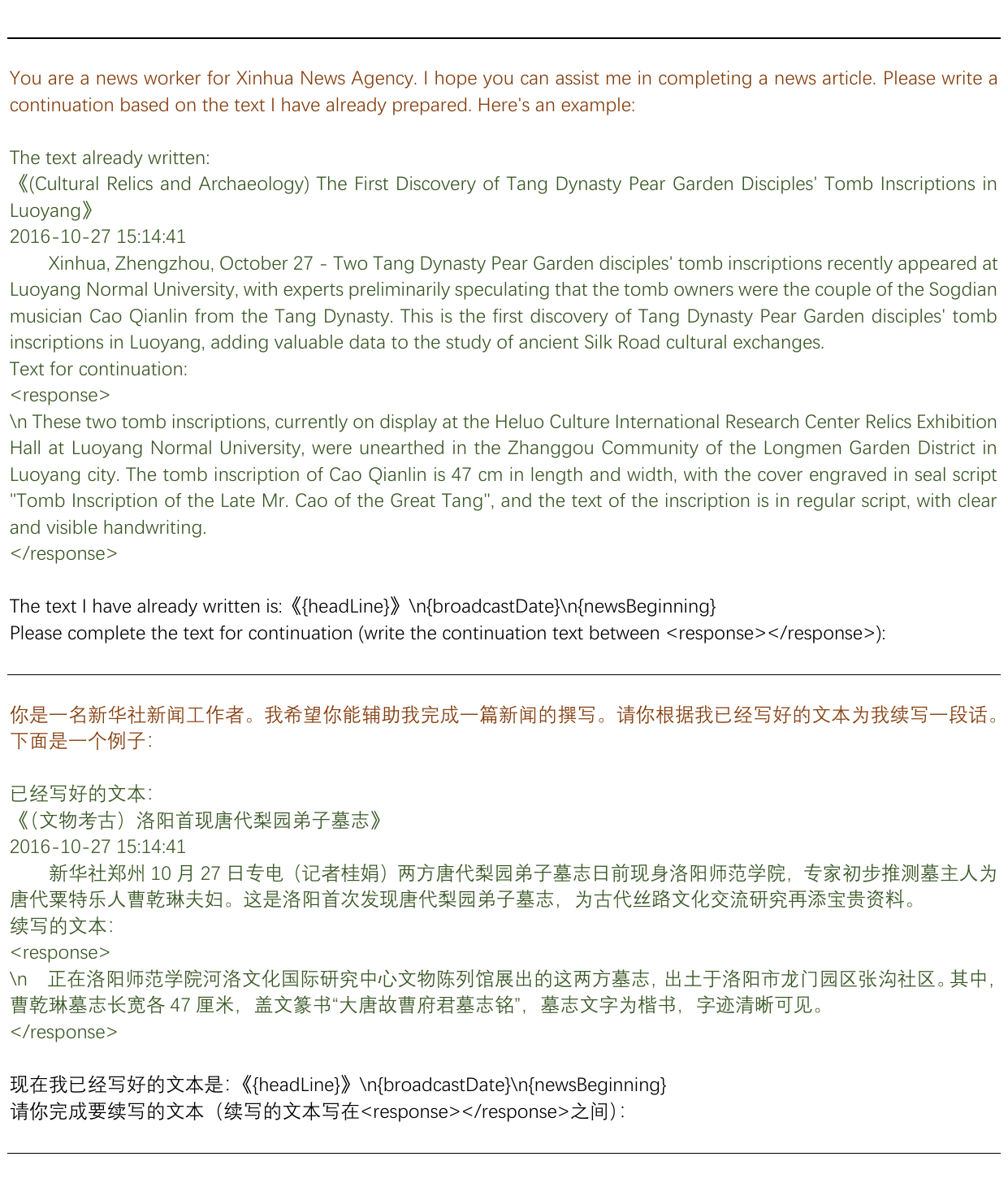}
    \caption{Prompt template for generative evaluation}
\end{figure}

\clearpage
\twocolumn
\section{Clarification of Imbalance of the Dataset in Term of Models} \label{apdx:imbalance}

\subsection{Imbalance Phenomenon}

The distribution of the final hallucination dataset in terms of five different LLMs used in the generation is shown in Table~\ref{tab:dataset_distribution}.

\begin{table}[h]
    \centering
    \begin{tabular}{lrr}
        \toprule
        \textbf{Generated by} & \textbf{Count} & \textbf{Share} \\
        \midrule
        Baichuan & 3425 & 66.62\% \\
        ChatGLM & 327 & 6.36\% \\
        Xinyu & 424 & 8.25\% \\
        InternLM & 487 & 9.47\% \\
        Qwen & 478 & 9.30\% \\
        \bottomrule
    \end{tabular}
    \caption{Dataset distribution by LLMs}
    \label{tab:dataset_distribution}
\end{table}

You may notice a disproportionately high representation of the Baichuan model, a discrepancy linked to an oversight at the outset. Initially, the generation of candidates did not fully leverage the available models, primarily because the Baichuan models exhibited superior instruction-following capabilities for data generation. Consequently, we solely utilized five instances of the Baichuan model for ranking to generate candidate data items. It was only later that four additional models were incorporated, employing a total of five distinct model instances for ranking to generate candidate data items. This approach resulted in a final dataset that lacks a relatively balanced distribution among the different models.

\subsection{Does the Imbalance Lead to Unreliable Outcomes?}

A quick answer is: No. Here are the justifications. 

We conducted a simple empirical study. Since the generated hallucinated texts are only used in selective and discriminative evaluations, if the imbalance significantly affects the experimental outcomes, it would only impact these two types of assessments. Fortunately, we have saved every piece of intermediate experimental results, allowing us to uniformly sample those results across models and re-aggregate the results to observe changes in key metrics like average accuracy. Specifically, we uniformly and randomly sampled 327 data points for each model and used a combined dataset of 327*5=1635 data points to re-aggregate the results. Below is a comparison of the experimental outcomes. Table~\ref{tab:original_results} and Table~\ref{tab:new_results} present the original results and the new results, respectively, while Table~\ref{tab:diff_original_new} displays the differences between them.

In analyzing the differences, we calculated three metrics to illustrate the magnitude of change. The average change in the three columns of accuracy metrics is approximately -0.0147, with a standard deviation of 0.0150. Furthermore, the average absolute change in their rankings is 0.606 (because only six pairs of closely ranked models undergo internal swaps). Lastly, upon separately examining the Baichuan model, which was anticipated to be most affected, we found its change to be not the greatest. Therefore, we can assert that the variance in model proportions has an insignificant impact on the outcomes.

Moreover, intuitively, since all these models are Chinese text generation models and the content they generate was also reviewed by our manual annotators, we eliminated some data items that significantly deviated the text type from the mean during annotation (for example, repetitions of previous content, or continuations that diverged from news to generating a reading comprehension question, etc.). This enhanced the overall consistency of the final dataset, making the task of distinguishing hallucinations in the experiment unrelated to the source of those hallucinations.

\begin{table*}[]
\begin{tabular}{llllllll}
\toprule
              & \multicolumn{3}{c}{\textbf{Discriminative-Keyword}} & \multicolumn{2}{c}{\textbf{Discriminative-Sentence}} & \multicolumn{2}{c}{\textbf{Selective}} \\
\cmidrule(lr){2-4} \cmidrule(lr){5-6} \cmidrule(lr){7-8}
              & avg.acc.     & avg.\#kws     & \#valid     & avg.acc.              & \#valid             & acc.          & \#valid       \\
\midrule
Aquila-34B    & 53.62\%      & 3.00          & 3719        & 49.86\%               & 5009                & 54.29\%       & 4319          \\
Baichuan2-13B & 51.63\%      & 3.13          & 4478        & 46.88\%               & 5047                & 50.23\%       & 5130          \\
Baichuan2-53B & 52.13\%      & 2.98          & 1656        & 50.81\%               & 1478                & 54.67\%       & 4443          \\
ChatGLM2-6B   & 50.80\%      & 3.10          & 4289        & 43.87\%               & 5130                & 43.59\%       & 5130          \\
GPT3.5-Turbo  & 53.72\%      & 3.08          & 4183        & 50.02\%               & 5039                & 49.03\%       & 5103          \\
GPT4-0613     & 70.04\%      & 3.07          & 4100        & 57.42\%               & 5024                & 55.20\%       & 5047          \\
GPT4-1106     & 69.48\%      & 3.10          & 4189        & 57.38\%               & 4903                & 60.35\%       & 4752         \\
InternLM-20B  & 50.92\%      & 3.10          & 4388        & 51.01\%               & 5130                & 49.43\%       & 5130          \\
Qwen-14B      & 52.86\%      & 3.13          & 4478        & 50.58\%               & 5130                & 54.74\%       & 5130          \\
Xinyu-7B      & 49.58\%      & 3.12          & 4451        & 48.66\%               & 5014                & 50.58\%       & 5130          \\
Xinyu2-70B     & 52.94\%      & 3.12          & 4482        & 55.04\%               & 5128                & 57.93\%       & 5129          \\
\bottomrule
\end{tabular}
\caption{Original Results}
\label{tab:original_results}
\end{table*}

\begin{table*}[]
\begin{tabular}{llllllll}
\toprule
              & \multicolumn{3}{c}{\textbf{Discriminative-Keyword}} & \multicolumn{2}{c}{\textbf{Discriminative-Sentence}} & \multicolumn{2}{c}{\textbf{Selective}} \\
\cmidrule(lr){2-4} \cmidrule(lr){5-6} \cmidrule(lr){7-8}
              & avg.acc.     & avg.\#kws     & \#valid     & avg.acc.              & \#valid             & acc.          & \#valid       \\
\midrule
Aquila-34B    & 54.17\%      & 3.18          & 1178        & 50.98\%               & 1582                & 57.34\%       & 1362          \\
Baichuan2-13B & 51.61\%      & 3.29          & 1398        & 50.34\%               & 1608                & 51.93\%       & 1629          \\
Baichuan2-53B & 52.68\%      & 3.06          & 525         & 51.46\%               & 479                 & 56.70\%       & 1432          \\
ChatGLM2-6B   & 51.27\%      & 3.27          & 1357        & 47.02\%               & 1629                & 46.04\%       & 1629          \\
GPT3.5-Turbo  & 54.38\%      & 3.23          & 1291        & 51.87\%               & 1601                & 50.06\%       & 1620          \\
GPT4-0613     & 70.03\%      & 3.23          & 1277        & 59.73\%               & 1593                & 58.79\%       & 1582          \\
GPT4-1106     & 68.24\%      & 3.25          & 1305        & 61.28\%               & 1547                & 65.34\%       & 1503         \\
InternLM-20B  & 51.06\%      & 3.23          & 1348        & 52.42\%               & 1629                & 53.53\%       & 1629          \\
Qwen-14B      & 53.84\%      & 3.29          & 1404        & 51.20\%               & 1629                & 53.96\%       & 1629          \\
Xinyu-7B      & 49.51\%      & 3.29          & 1389        & 48.74\%               & 1582                & 50.58\%       & 1629          \\
Xinyu2-70B     & 54.30\%      & 3.29          & 1402        & 58.24\%               & 1627                & 59.28\%       & 1628          \\
\bottomrule
\end{tabular}
\caption{New Results}
\label{tab:new_results}
\end{table*}

\begin{table*}[]
\begin{tabular}{llllllll}
\toprule
              & \multicolumn{3}{c}{\textbf{Discriminative-Keyword}} & \multicolumn{2}{c}{\textbf{Discriminative-Sentence}} & \multicolumn{2}{c}{\textbf{Selective}} \\
\cmidrule(lr){2-4} \cmidrule(lr){5-6} \cmidrule(lr){7-8}
              & avg. acc.     & avg. \#kws    & \#valid    & avg. acc.             & \#valid             & acc.          & \#valid       \\
\midrule
Aquila-34B    & -0.55\%       & -0.18         & 2541       & -1.12\%               & 3427                & -3.05\%       & 2957          \\
Baichuan2-13B & 0.02\%        & -0.16         & 3080       & -3.46\%               & 3439                & -1.70\%       & 3501          \\
Baichuan2-53B & -0.55\%       & -0.08         & 1131       & -0.65\%               & 999                 & -2.03\%       & 3011          \\
ChatGLM2-6B   & -0.48\%       & -0.17         & 2932       & -3.15\%               & 3501                & -2.45\%       & 3501          \\
GPT3.5-Turbo  & -0.67\%       & -0.15         & 2892       & -1.85\%               & 3438                & -1.03\%       & 3483          \\
GPT4-0613     & 0.01\%        & -0.16         & 2823       & -2.31\%               & 3431                & -3.59\%       & 3465          \\
GPT4-1106     & 1.24\%        & -0.14         & 2884       & -3.90\%               & 3356                & -4.98\%       & 3249         \\
InternLM-20B  & -0.14\%       & -0.13         & 3040       & -1.41\%               & 3501                & -4.10\%       & 3501          \\
Qwen-14B      & -0.98\%       & -0.17         & 3074       & -0.62\%               & 3501                & 0.78\%        & 3501          \\
Xinyu-7B      & 0.07\%        & -0.17         & 3062       & -0.07\%               & 3432                & 0.00\%        & 3501          \\
Xinyu2-70B     & -1.36\%       & -0.16         & 3080       & -3.20\%               & 3501                & -1.35\%       & 3501          \\
\bottomrule
\end{tabular}
\caption{Difference (Original - New)}
\label{tab:diff_original_new}
\end{table*}

\end{document}